\pdfoutput=1

\documentclass[11pt]{article}
\usepackage{geometry}
\usepackage{hyperref}
\usepackage{amsmath}
\usepackage{enumitem}
\usepackage{pifont}
\usepackage[utf8]{inputenc} 
\usepackage{booktabs}  
\usepackage{longtable}
\usepackage{tabularx}
\usepackage{float}
\usepackage{graphicx}    
\usepackage{multirow}    
\usepackage{caption}     
\usepackage{amssymb}
\usepackage{authblk}
\usepackage{xcolor}
\usepackage{mdframed}
\usepackage{fancyvrb}
\usepackage{float}

\usepackage[preprint]{acl}

\usepackage{times}
\usepackage{latexsym}
\usepackage[T1]{fontenc}
\usepackage{microtype}
\usepackage{inconsolata}
\usepackage{amsmath}  
\newcommand{\cmark}{\ding{51}} 
\newcommand{\xmark}{\ding{55}} 
\newcommand{\equalcontrib}{\textsuperscript{\textdagger}}  
\newcommand{\correspondauthor}{\textsuperscript{*}}
\usepackage{float}  

\usepackage{xcolor}
\usepackage{mdframed}

\title{ACEBench: Who Wins the Match Point in Tool Usage?}

\author{
  \textbf{Chen Chen}\textsuperscript{1}\equalcontrib, 
  \textbf{Xinlong Hao}\textsuperscript{2}\equalcontrib, 
  \textbf{Weiwen Liu}\textsuperscript{3}\correspondauthor,
  \textbf{Xu Huang}\textsuperscript{1}, 
  \textbf{Xingshan Zeng}\textsuperscript{2}, \\
  \textbf{Shuai Yu}\textsuperscript{2}, 
  \textbf{Dexun Li}\textsuperscript{2}, 
  \textbf{Shuai Wang}\textsuperscript{2}, 
  \textbf{Weinan Gan}\textsuperscript{2}, 
  \textbf{Yuefeng Huang}\textsuperscript{1}, \\
  \textbf{Wulong Liu}\textsuperscript{2}, 
  \textbf{Xinzhi Wang}\textsuperscript{2}, 
   \textbf{Defu Lian}\textsuperscript{1},
   \textbf{Baoqun Yin}\textsuperscript{1},
  \textbf{Yasheng Wang}\textsuperscript{2}\correspondauthor,
   \textbf{Wu Liu}\textsuperscript{1}\correspondauthor,

\\
\\
 \textsuperscript{1}University of Science and Technology of China,
 \textsuperscript{2}Huawei Noah's Ark Lab,\\
 \textsuperscript{3}Shanghai Jiao Tong University,

\\
 \small{
 \href{mailto:email@domain}{chenchen0318@mail.ustc.edu.cn}
 \href{mailto:email@domain}{haoxinlong@huawei.com         
 }
 }}

\begin{document}

\maketitle

\begin{abstract}

Large Language Models (LLMs) have demonstrated significant potential in decision-making and reasoning, particularly when integrated with various tools to effectively solve complex problems. However, existing benchmarks for evaluating LLMs' tool usage face several limitations: (1) limited evaluation scenarios, often lacking assessments in real multi-turn dialogue contexts; (2) narrow evaluation dimensions, with insufficient detailed assessments of how LLMs use tools; and (3) reliance on LLMs or real API executions for evaluation, which introduces significant overhead. To address these challenges, we introduce ACEBench, a comprehensive benchmark for assessing tool usage in LLMs. ACEBench categorizes data into three primary types based on evaluation methodology: Normal, Special, and Agent. "Normal" evaluates tool usage in basic scenarios; "Special" evaluates tool usage in situations with ambiguous or incomplete instructions; "Agent" evaluates tool usage through multi-agent interactions to simulate real-world, multi-turn dialogues. We conducted extensive experiments using ACEBench, analyzing various LLMs in-depth and providing a more granular examination of error causes across different data types.  
\end{abstract}

{
\renewcommand{\thefootnote}{\textdagger}
\footnotetext[1]{Equal Contributions. Work was done during an internship at Huawei Noah's Ark Lab. \textsuperscript{*}Corresponding authors.}
\renewcommand{\thefootnote}{*}
\footnotetext[3]{The code are already publicly available at \href{https://github.com/chenchen0103/ACEBench}{GitHub}.}
\renewcommand{\thefootnote}{\textdagger}
}

\section{Introduction}

Large Language Models (LLMs), such as GPT-4~\cite{gpt4}, have demonstrated exceptional performance across numerous natural language processing  tasks~\cite{llmsurvey,toolsurvey,survey2}. 
Studies have shown that incorporating tools can significantly expand LLM capabilities, particularly in specialized domains such as mathematics~\cite{mathsensei,math_bulusu2024mathviz,math_gou2023tora,math_veerendranath2024calc}, programming~\cite{xu2024core}, and reasoning~\cite{reason_chen2022program,reason_shao2022chaining,reason_suris2023vipergpt,reason_yang2023mm}. On one hand, integrating tools into LLMs can enhance capabilities in multiple domains, for example, ToolTransformer~\cite{tooltransformer} enhances the ability of LLMs to solve complex problems by utilizing tools. On the other hand, adopting a tool usage paradigm can improve the robustness of the response and the transparency of the generation, thus increasing the explainability and trust of usersusers~\cite{tooltransformer}, as well as improving the system's adaptability. As this field continues to evolve, it is essential to comprehensively evaluate all aspects of tool usage, particularly in complex scenarios.  


\begin{table*}[th]
\centering
\renewcommand{\arraystretch}{1} 
\setlength{\tabcolsep}{6pt} 
\caption{Comparison of benchmarks across different evaluation criteria. "LLM-Free" refers to result evaluation without relying on LLMs. "Robustness" refers to incomplete or unclear user instructions. "Interactiveness" refers to the dynamic interaction between the model and the environment. "Atomic-Level" refers to analyzing from the atomic-level capabilities. "Personalization” refers to the inclusion of personal likes.}
\resizebox{\textwidth}{!}{
\footnotesize 
\begin{tabular}{lccccc}
\toprule
\textbf{Benchmark} & \textbf{LLM-Free} & \textbf{Robustness} & \textbf{Interactiveness} & \textbf{Atomic-Level} & \textbf{Personalization} \\
\midrule
MetaTool~\cite{metatool}         & \cmark & \xmark & \xmark & \xmark & \xmark \\
API-Bank~\cite{apibank}         & \cmark & \xmark & \xmark & \xmark & \xmark \\
Stable ToolBench~\cite{stabletoolbench} & \xmark & \xmark & \xmark & \xmark & \xmark \\
BFCL~\cite{bfcl}             & \cmark & \cmark & \xmark & \xmark & \xmark \\
$\tau$-Bench~\cite{taubench}         & \cmark & \xmark & \cmark & \xmark & \xmark \\
HammerBench~\cite{hammerbench}      & \xmark & \cmark & \xmark & \xmark & \xmark \\
ACEBench (Ours)  & \cmark & \cmark & \cmark & \cmark & \cmark \\
\bottomrule
\end{tabular}
}
\label{tab:benchmark_comparison}
\end{table*}

While several studies have focused on evaluating tool usage~\cite{bfcl,stabletoolbench,hammerbench,toolllm,mtubench,toolqa,toolsandbox}, there are still some shortcomings in the existing tool-use benchmarks. Firstly, existing benchmarks lack multi-turn dialogue evaluation in real-world scenarios. For example, the multi-turn dialogues in BFCL~\cite{bfcl} and HammerBench ~\cite{hammerbench} are composed of predefined fixed content combinations. Secondly, current tool-use benchmarks~\cite{toolllm,stabletoolbench,metatool,apibank} lack fine-grained evaluation and personalized data assessment. 
Additionally, existing benchmarks~\cite{toolllm,stabletoolbench,mtubench} ignore the assessment of special cases, or the evaluation methods are simplistic~\cite{bfcl}, as user instructions in real life are not always perfect\cite{nosiy_toolbench}. The model’s ability to recognize and handle these issues is also crucial for evaluation. Lastly, evaluation costs are high ~\cite{toolllm,stabletoolbench}, as many studies rely on advanced large models for evaluation.

To address these shortcomings, we propose ACEBench, a comprehensive tool-use benchmark that includes the following categories:

\noindent\textbf{Normal.} Consists of fixed question-answer pairs and encompasses a variety of scenarios, including single-turn dialogues, multi-turn dialogues, and personalized scenario data. It also includes evaluations of atomic-level capabilities.

\noindent\textbf{Special.} Includes imperfect instructions, such as instructions containing incomplete parameters, incorrectly formatted parameters, or questions irrelevant to the capabilities of the candidate functions.

\noindent\textbf{Agent.} Encompasses real-world scenarios, abstracted to construct multi-turn, multi-step tool invocation scenarios, divided into multi-turn and multi-step cases depending on whether the user participates in the dialogue process.

The three categories above cover most of the tool usage scenarios for LLMs, and detailed explanations of each category can be found in Appendix~\ref{sec:data_description}. Our main contributions are as follows:
\begin{itemize}[leftmargin=*, noitemsep]
    \item \textbf{Comprehensive Benchmark Evaluation.} We propose a comprehensive benchmark for evaluating LLMs' tool usage, covering various scenarios, including more fine-grained evaluation perspectives and assessments under imperfect instructions and providing more stable evaluation metrics.
    \vspace{0.5\baselineskip} 
    \item \textbf{Sandbox Environment and Automated Evaluation System.} We build an end-to-end automated evaluation system and develop a sandbox environment construction scheme for multi-turn, multi-step tool invocation based on real-world scenario abstraction.
    \vspace{0.5\baselineskip} 
    \item \textbf{Extensive Experimental Validation.} Through extensive experiments, we demonstrate our benchmark provides a more comprehensive analysis with greater distinction, offering a clearer evaluation of LLMs' tool usage.
\end{itemize}


\begin{figure*}[t]
  \centering
  \includegraphics[width=\textwidth]{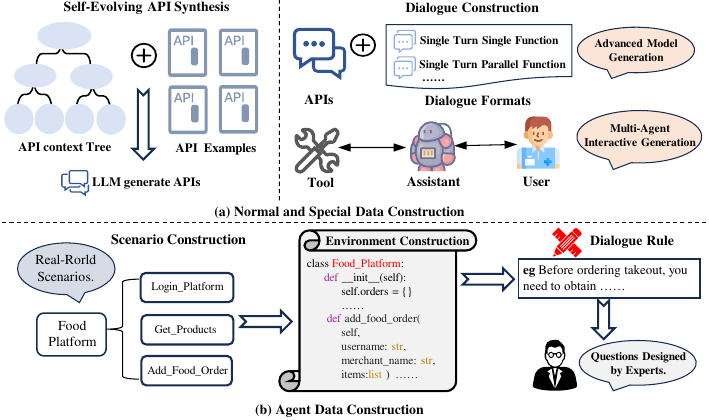}
  \caption{Dataset construction pipeline. (a) Normal and Special data construction: API synthesis module (left), Dialogue generation module (right). (b) Agent Data Construction: include scenario construction, environment construction and question design.}
  \label{fig:data_pipeline}
\end{figure*}

\section{Related Works}
The emerging trend of leveraging LLMs' tool-use capabilities in real-world applications underscores the need for comprehensive evaluations of their performance and effectiveness.
Despite recent advancements, existing benchmarks for evaluating the tool-use capabilities of LLMs still have significant limitations

Stable ToolBench~\cite{stabletoolbench} addresses the issue of unstable external APIs by employing a virtual API server, but its dependence on large models for evaluation results in high costs and scalability challenges. 
BFCL~\cite{bfcl} introduces a benchmark for tool use in multi-turn dialogue scenarios. Yet, it assembles dialogues from fixed content, failing to capture the dynamic and adaptive nature of real-world interactions. 
Similarly, $\tau$-Bench~\cite{taubench} evaluates language agents' ability to engage with human users while adhering to domain-specific rules. Still, its narrow focus on just two scenarios limits its generalizability across diverse tasks.
HammerBench~\cite{hammerbench} improves upon this by incorporating datasets derived from popular mobile applications and merging dialogues to simulate typical question-answer trajectories. However, like BFCL, its multi-turn dialogues are simplistic concatenations of pre-defined content, which do not reflect the complexities of real-world conversational dynamics. 
In addition, some benchmarks~\cite{toolllm,stabletoolbench} rely on large language models (LLMs) for result evaluation, leading to high costs and unstable operations.

In contrast, our work addresses these limitations by expanding the scope of evaluation to encompass a broader range of tool usage scenarios. 
We propose a framework that simulates realistic multi-turn dialogue processes and enables end-to-end automated assessment, thereby reducing evaluation costs and improving scalability. 
A comparative analysis of ACEBench against recent benchmarks, as shown in Table~\ref{tab:benchmark_comparison}, demonstrates its effectiveness in overcoming these challenges.


\section{ACEBench}

\subsection{Dataset}
We constructed two linguistically parallel versions of the dataset (Chinese and English), ensuring equal distribution of data types between them. The final dataset comprises 2,000 annotated entries.

\subsubsection{Data Construction}  
\noindent\textbf The Normal and Special data are automatically generated by LLMs, whereas the Agent data is constructed by experts. Creation details for some data are provided in Appendix Section~\ref{sec:construct_data}.

\noindent\textbf{Normal and Special Data Construction. } We employ a fully automated LLM-based generation pipeline specifically designed for Normal and Special Data , as illustrated in Figure~\ref{fig:data_pipeline}(a).

\noindent\textbf{(1) API Synthesis.}
We use real APIs from various real-world scenarios as reference during construction to enhance authenticity. To ensure the stability of the data, we use synthetic APIs to construct the evaluation dataset, referencing real-world APIs as a guide. We employ a self-evolution approach by building a hierarchical API context tree to ensure the generated APIs cover a wide range of domains and functionalities~\cite{toolace}. Initially, we extract relevant information from technical documents to guide the API generation. As the process progresses, the context tree is gradually expanded, ultimately ensuring the depth and breadth of the generated APIs. 

\noindent\textbf{(2) Dialogue Construction.}
We use two different dialogue generation pipelines built on the constructed API pool from which three to six candidate APIs are selected for each evaluation instance. For most cases, APIs are chosen randomly. However, for instances requiring specific functionality (e.g., similar APIs or multi-turn scenarios), advanced methods, including graph-based sampling~\cite{toolflow}, are used. Simple cases or those with predefined functionality use a template-based generation, where a single generator produces dialogues to ensure consistency. We employ a multi-agent dialogue pipeline for more complex scenarios, where three agents (user, assistant, and tool) to simulate real-world interactions. Both pipelines are supported by carefully hand-crafted examples to ensure comprehensive coverage and diversity.

\noindent\textbf{Agent Data Construction.} We implement a carefully curated human-expert construction framework specifically tailored for Agent Data generation, as shown in Figure~\ref{fig:data_pipeline}(b). 

\noindent\textbf{(1) Scenario Construction.} Through systematic abstraction of real-world interaction scenarios (such as food delivery services and telecommunication operations), we design functional modules with well-defined business semantics and specify each scenario's core state variables (e.g., order status, account balance) and intrinsic property sets.

\noindent\textbf{(2) Sandbox Environment Construction.} We constructed an isolated sandbox environment with three core components: standardized functional interfaces with well-defined input/output specifications and preconditions, a dynamic attribute management system for real-time state transition monitoring, and an execution monitoring module that logs invocation processes.

\noindent\textbf{(3) Question Design.} Based on predefined multi-turn dialog specifications tailored to different scenarios, domain experts systematically crafted the conversational questions through an iterative annotation process.


\begin{figure}[t]
    \centering
    \includegraphics[width=0.9\linewidth]{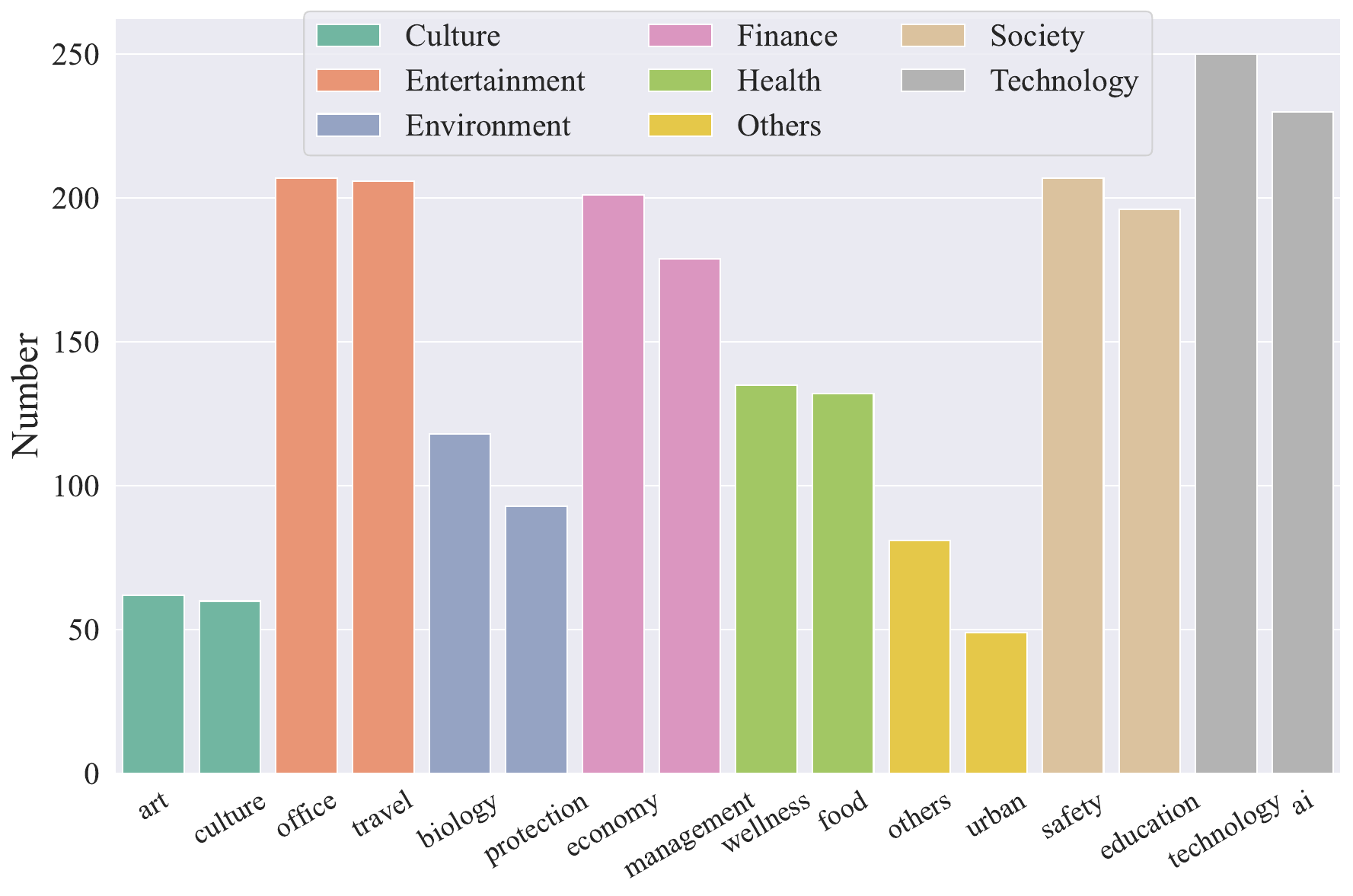}
    \caption{Distribution of APIs in terms of domains (Top 2 subcategories for each category).}
    \label{fig:api_domain_dist}
\end{figure}

\subsubsection{Multi-Stage Data Verification}

To address issues like mismatched answers or ambiguous criteria, we have implemented a multi-stage verification process.

\noindent\textbf{Automated Quality Inspection. } The data first undergoes a rule-based quality inspection module, which evaluates four dimensions: clarity of API definitions, executability of function calls, accuracy of dialogues, and consistency of data samples, effectively filtering out formatting and spelling errors. Next, the data enters the model-based quality verification module, which uses LLMs to detect semantic errors, employing a voting mechanism to ensure consistency in evaluation.

\noindent\textbf{Human Quality Inspection. }In the initial evaluation, the dataset remaining after automated quality inspection is assessed by three LLMs to assist human experts in data screening. Valid data is retained, while potentially problematic data is placed in the error candidate pool. These flagged entries undergo a two-step expert review process, where two experts independently assess and suggest modifications, and a third expert consolidates feedback, revising problem statements, API definitions, and answers. The revised data is re-evaluated and manually verified, and three rounds of optimization are performed to ensure a high-quality dataset.


\subsubsection{Data Analysis}

To demonstrate the breadth and comprehensiveness of ACEBench, we provide a detailed analysis of its test case distributions. Specific examples of each data type can be found in Appendix~\ref{sec:data_example}.

\noindent\textbf{Domain of APIs}.
The ACEBench API boasts a comprehensive coverage of 8 major domains and 68 sub-domains, spanning various aspects of daily life, including technology, finance, entertainment, society, health, culture, environment, and others. It offers a rich collection of 4,538 APIs in both Chinese and English. The distribution of these APIs is visualized in the accompanying Figure~\ref{fig:api_domain_dist}.

\begin{figure}[t]
    \centering
    \includegraphics[width=0.69\linewidth]{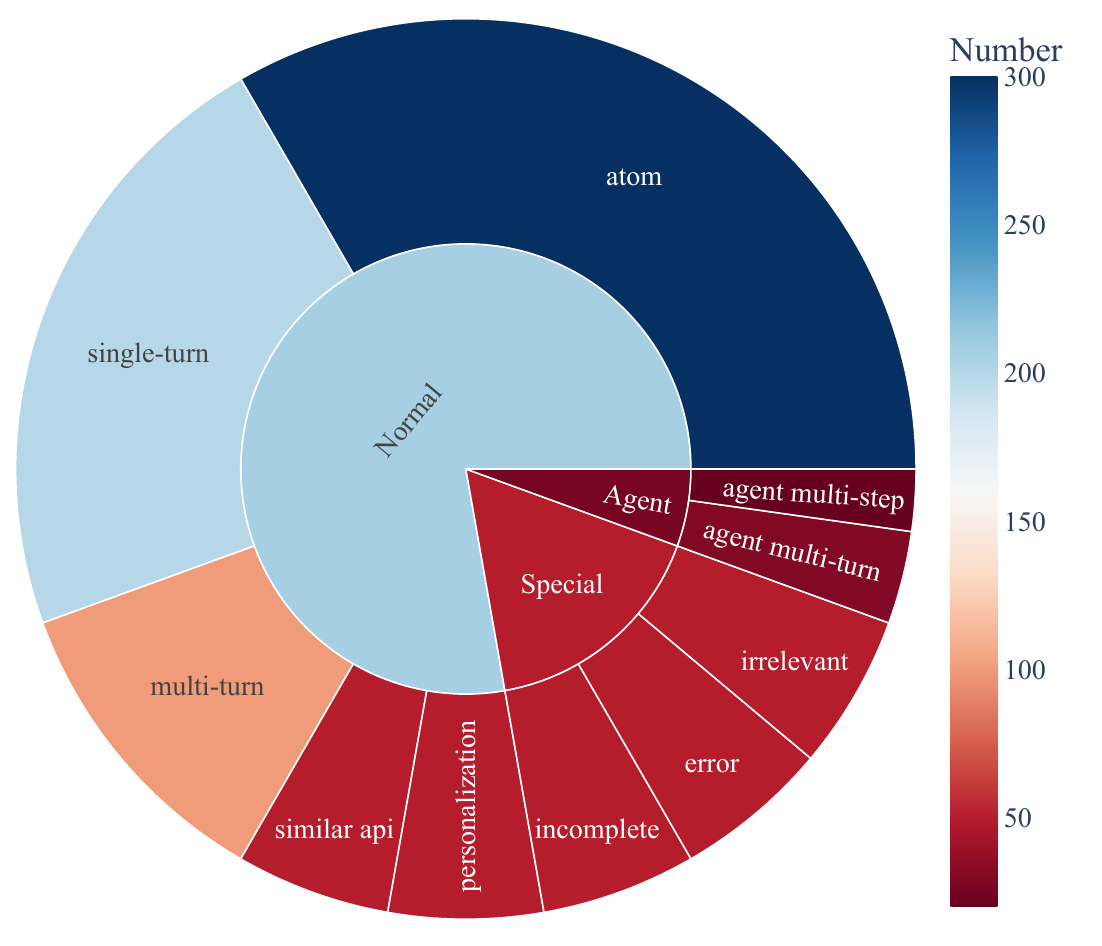}
    \caption{Visualization of the data composition of ACEBench.}
    \label{fig:data_composite}
\end{figure}

\noindent\textbf{Data Composition.} 
ACEBench consists of three categories of test samples: Normal, Agent, and Special, where each category is divided into several subcategories. The data composition is visualized in Figure~\ref{fig:data_composite}, demonstrating a comprehensive coverage of tool-use capabilities, from simple single-turn tool invocations to complex multi-turn interactions involving users and environments. They include scenarios requiring multiple steps and interactions with the environment, as well as cases where tool calls are infeasible.

\begin{figure}[t]
    \centering
    \includegraphics[width=1\linewidth]{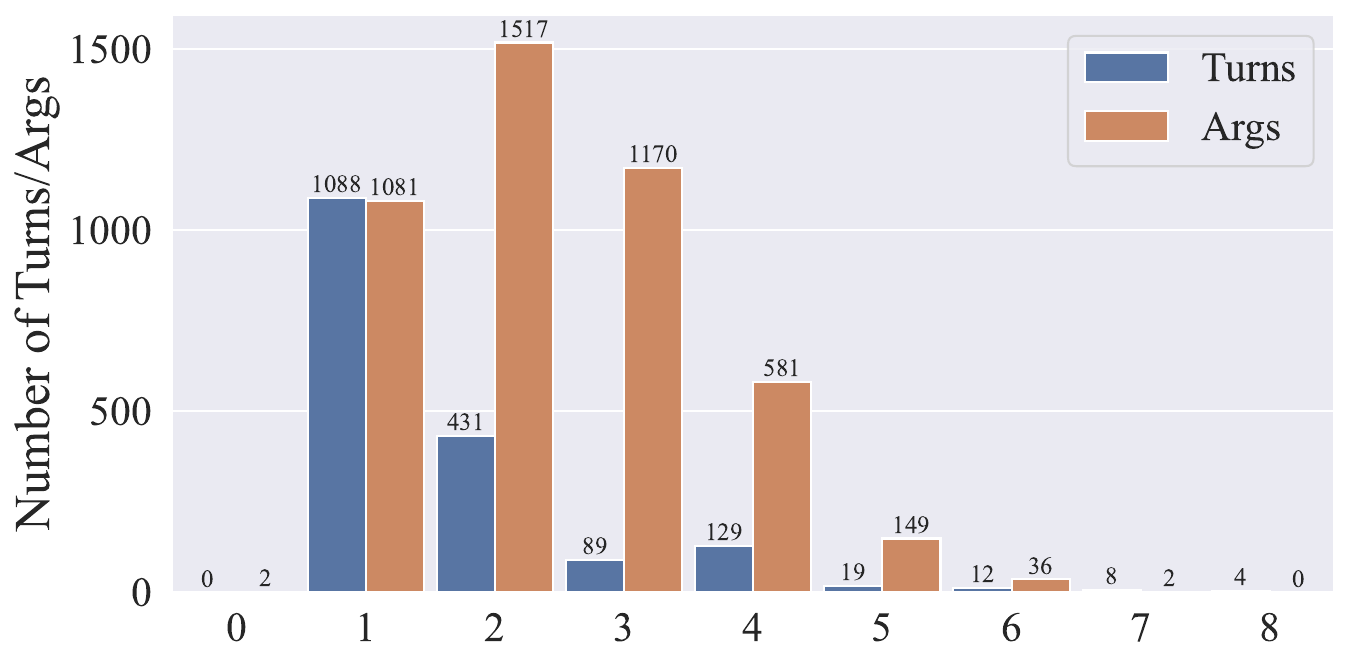}
    \caption{Distribution of dialogue turns and API argument numbers.}
    \label{fig:turn_and_arg}
\end{figure}

\noindent\textbf{Number of turns and arguments.} 
The test data in ACEBench covers a wide range of complexities. Specifically, we statistically analyzed the number of dialogue turns and the number of arguments in the called apis, which are visualized in Figure~\ref{fig:turn_and_arg}. The results show that the number of dialogue turns ranges from 1 to 8, encompassing most real-world scenarios. 
These samples with varying numbers of turns and arguments further form a test suite that covers a broader range of difficulties.

\begin{table*}[t]
    \centering
    \caption{Comprehensive evaluation of different models on ACEBench for Chinese and English combined (\%).}
    \resizebox{\textwidth}{!}{%
    \begin{tabular}{lcccccccccc}
        \toprule
        \multirow{2}{*}{\textbf{Model}} & 
        \multicolumn{6}{c}{\textbf{Normal}} & 
        \multicolumn{1}{c}{\multirow{2}{*}{\textbf{Special}}} & 
        \multicolumn{1}{c}{\multirow{2}{*}{\textbf{Agent}}} & 
        \multicolumn{1}{c}{\multirow{2}{*}{\textbf{Overall}}} \\
        \cmidrule(lr){2-7}
        & \textbf{Atom} & \textbf{Single-Turn} & \textbf{Multi-Turn} & \textbf{Similar API} & \textbf{Preference} & \textbf{Summary} & & & \\
        \midrule
        \multicolumn{10}{c}{\textbf{Closed-Source Large Language Models}} \\
        \midrule
        GPT-4o & 93.4 & 84.5 & 77.0 & 85.0 & 83.0 & 87.6 & 93.0 & 63.8 & \textbf{85.4} \\
        GPT-4-Turbo & 93.2 & 84.8 & 77.5 & 86.0 & 86.0 & 88.0 & 86.7 & 67.5 & \textbf{84.5} \\
        Qwen-Max & 91.2 & 80.5 & 68.0 & 83.0 & 83.0 & 84.2 & 74.0 & 64.3 & \textbf{78.4} \\
        GPT-4o-Mini & 86.5 & 76.0 & 66.5 & 77.0 & 78.0 & 79.9 & 79.0 & 33.3 & \textbf{72.5 }\\
        Gemini-1.5-Pro & 84.5 & 76.8 & 64.5 & 80.0 & 78.0 & 79.0 & 78.7 & 25.5 & \textbf{70.7} \\
        Claude-3-5-Sonnet & 76.9 & 72.5 & 62.5 & 71.0 & 72.0 & 72.9 & 77.4 & 39.5 & \textbf{68.9} \\
        Doubao-Pro-32k & 79.8 & 55.5 & 58.0 & 76.0 & 66.0 & 70.7 & 55.0 & 25.0 & \textbf{59.4} \\
        \midrule
        \multicolumn{10}{c}{\textbf{Open-Source Large Language Models}} \\
        \midrule
        Qwen2.5-Coder-32B-Instruct & 90.2 & 81.0 & 71.0 & 83.0 & 81.0 & 84.1 & 80.7 & 60.8 & \textbf{79.6} \\
        DeepSeek-V3 & 91.5 & 84.0 & 77.0 & 83.0 & 83.0 & 86.5 & 73.0 & 34.5 & \textbf{74.8} \\
        Qwen2.5-72B-Instruct & 86.8 & 80.3 & 69.5 & 83.0 & 81.0 & 82.1 & 75.7 & 45.0 & \textbf{74.7} \\
        Llama-3.1-70B-Instruct & 82.5 & 68.3 & 63.5 & 79.0 & 68.0 & 75.5 & 38.3 & 42.3 & \textbf{60.4} \\
        Qwen2.5-7B-Instruct & 76.0 & 60.3 & 58.5 & 72.0 & 67.0 & 69.4 & 47.0 & 13.8 & \textbf{54.8} \\
        DeepSeek-Coder-V2-Lite-Instruct & 75.2 & 57.8 & 46.5 & 72.0 & 65.0 & 66.4 & 40.3 & 2.0 & \textbf{49.5} \\
        Qwen2.5-Coder-7B-Instruct & 76.0 & 63.8 & 57.5 & 74.0 & 68.0 & 70.1 & 22.3 & 15.5 & \textbf{48.9} \\
        Watt-Tool-8B & 85.7 & 69.3 & 55.5 & 79.0 & 64.0 & 75.6 & 6.0 & 2.8 & \textbf{45.7} \\
        Hammer2.1-7B & 73.7 & 57.5 & 40.0 & 62.0 & 55.0 & 62.8 & 14.7 & 16.8 & \textbf{42.9} \\
        Llama-3.1-8B-Instruct & 51.9 & 39.8 & 28.0 & 66.0 & 46.0 & 46.6 & 21.0 & 5.3 & \textbf{33.4} \\
        Phi-3-Mini-128k-Instruct & 57.2 & 39.3 & 23.0 & 58.0 & 32.0 & 46.5 & 18.7 & 0.8 & \textbf{32.0} \\
        xLAM-7B-r & 43.5 & 22.0 & 19.0 & 61.0 & 0.0 & 33.7 & 2.7 & 8.8 & \textbf{21.6} \\
        Llama-3.2-3B-Instruct & 38.7 & 15.3 & 9.0 & 42.0 & 32.0 & 29.6 & 9.4 & 0.0 & \textbf{19.6} \\
        Hammer2.1-3B & 22.4 & 11.5 & 3.5 & 40.0 & 20.0 & 18.7 & 1.0 & 1.5 & \textbf{11.3} \\
        \bottomrule
    \end{tabular}%
    }
    \label{tab:results_all}
\end{table*}

\begin{figure*}[thb]
  \centering
  \includegraphics[width=\textwidth]{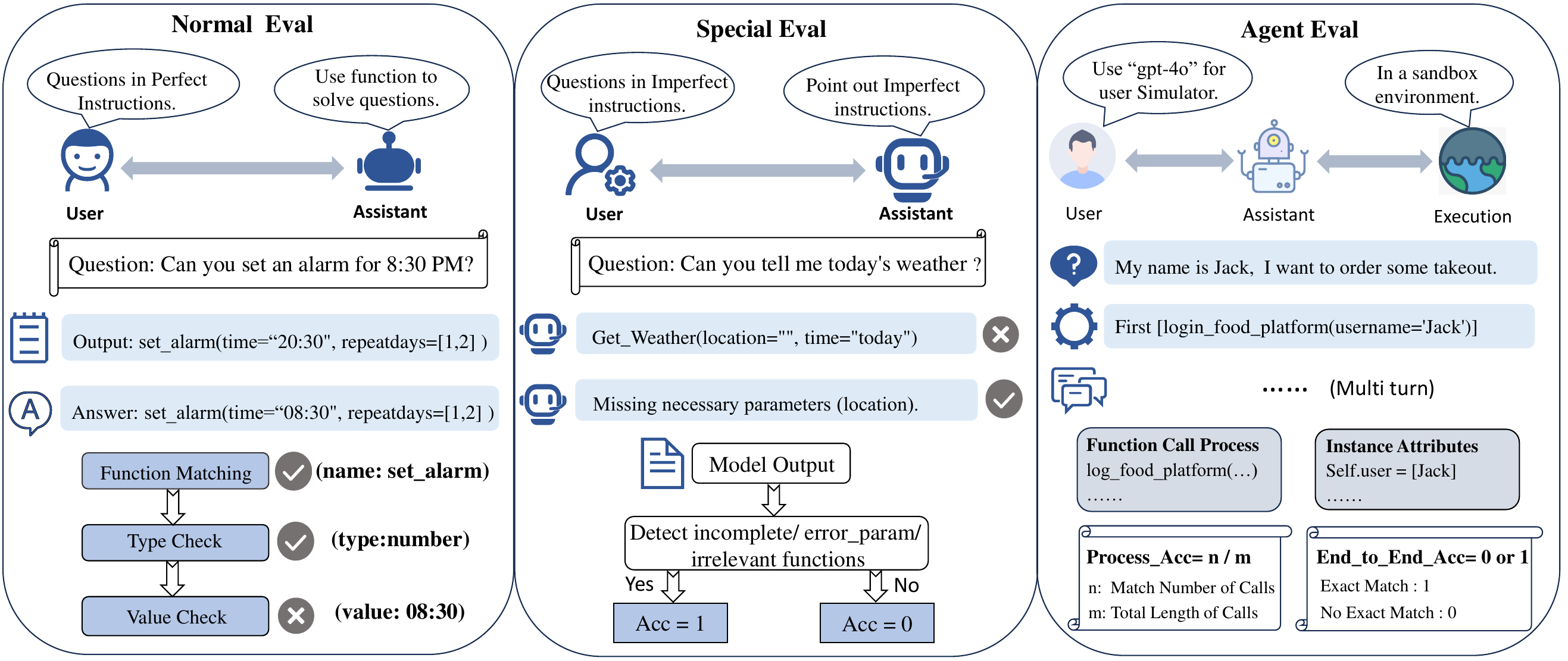}
  \caption{Overview of evaluation process: The left represents 'Normal' evaluation: AST-based function and parameter verification. The middle illustrates 'Special' evaluation: Imperfect instruction defect diagnosis. The right shows 'Agent' evaluation: State transition analysis via user-model interaction}
  \label{fig:eval}
\end{figure*}

\begin{table}[hb]
\centering
\caption{The Accuracy of Models on Special Data}
\renewcommand{\arraystretch}{1} 
\resizebox{\linewidth}{!}{ 
\begin{tabular}{lcccc}
\toprule
\textbf{Model}  & \multicolumn{1}{c}{\textbf{Incomplete}} & \multicolumn{1}{c}{\textbf{Error}} & \multicolumn{1}{c}{\textbf{Irrelevant}} \\
\midrule
 Llama-3.1-8B-Instruct & 29.0 & 20.0 & 14.0  \\
 Watt-Tool-8B & 7.0 & 1.0 & 10.0  \\
 \midrule
 Qwen2.5-7B-Instruct & 26.0 & 36.0 & 79.0  \\
 xLAM-7B-r & 1.0 & 3.0 & 4.0  \\
 \midrule
 Llama-3.2-3B-Instruct & 13.0 & 12.0 & 3.0  \\
 Hammer2.1-3B & 0.0 & 3.0 & 0.0  \\
\bottomrule
\end{tabular}
}
\label{tab:special}
\end{table}

\subsection{Eval}
\label{sec:eval}

In this section, we introduce the evaluation process, and the inference prompt for evaluation can be found in Appendix~\ref{sec:inference_prompt}.

\subsubsection{Normal Evaluation}  

As shown in the left part of Figure~\ref{fig:eval}, we evaluate Normal Data by comparing the model's function call output with the ground truth using AST parsing. For cases with multiple valid answers, we employ a candidate answer pool where matching any candidate constitutes correctness. Evaluation uses Accuracy metric (1=full match, 0=mismatch).

\subsubsection{Special Evaluation}  

As illustrated in the middle section of Figure~\ref{fig:eval}, the evaluation of Special Data primarily assesses the model's capability in problem identification. Specifically, the model must: (1) detect and alert missing parameters, (2) accurately locate erroneous parameters, and (3) recognize task-function mismatches. For each case, Accuracy is scored as 1 if correctly identified, otherwise 0.

\subsubsection{Agent Evaluation}
As shown in the right part of Figure~\ref{fig:eval}, we evaluate the agent's capabilities by assessing the model's proficiency in utilizing tools during human-agent interactions, employing gpt-4o as a user simulator for testing purposes. There are two evaluation metrics:

\noindent\textbf{End-to-End Accuracy} is evaluated by comparing the instance attributes of the corresponding class with the target. If all attributes match exactly, the Accuracy is 1; otherwise, the Accuracy is 0.

\noindent\textbf{Process Accuracy} is determined by the consistency between the actual function call process and the ideal process. It is expressed as \( \frac{n}{m} \), where \( m \) represents the ideal function call process, and \( n \) represents the degree of match between the actual and ideal processes.


\subsubsection{Overall Accuracy}

The Overall Accuracy is computed as a weighted sum of the accuracies for the Normal, Special, and Agent data types, where the weights are determined by the square roots of their respective sample sizes. The details can be found in Appendix~\ref{sec:formula}.

\section{Experiments}

In this section, we present a comprehensive set of experiments designed to evaluate the performance of LLMs on ACEBench.

\noindent\textbf{Experimental Setup.} In our evaluation, we examine seven closed-source LLMs, including the GPT-4 series~\cite{gpt4}, Qwen-Max~\cite{qwen2.5}, Gemini-1.5-Pro~\cite{gemini1.5}, Claude-3.5-Sonnet~\cite{claude-3.5}, and Doubao-Pro-32K~\cite{doubao-pro-32k}. Additionally, a wide range of open-source language models are assessed, such as the Qwen2.5 series~\cite{qwen2.5}, Llama3 series~\cite{llama3}, Phi-3-Mini~\cite{phi3}, Deepseek-V3\cite{deepseek-v3}, and DeepSeek-Coder-V2~\cite{DeepSeek-Coder-V2}. Furthermore, four tool-learning-enhanced models were evaluated: Hammer2.1-3B, Hammer2.1-7B~\cite{hammer}, xLAM-7B-r~\cite{xlam}, and Watt-Tool-8B~\cite{watt-tool}.

We additionally report the results of Kimi-K2~\cite{kimik2} and LoopTool-8B~\cite{looptool} on the English dataset.

\subsection{Main results and analysis}

The comprehensive experimental results for the Chinese and English datasets are presented in Table~\ref{tab:results_all}, with detailed results for each language provided in Appendix~\ref{sec:detailed_results}. We can draw the following important conclusions:

\vspace{0.5em}

\noindent \textbf{General Conclusion on Model Performance.}
The overall best performance remains dominated by closed-source models, such as the GPT-4 series. However, the performance gap between certain open-source models, such as Qwen2.5-Coder-32B-Instruct, Qwen2.5-72B-Instruct and DeepSeek-V3, and their closed-source counterparts is progressively narrowing. This trend suggests that open-source models are steadily catching up to closed-source models, driven by advancements in model architecture and training methodologies.

\noindent\textbf{Loss of Generalization in Fine-Tuned Models.} As shown in Figure~\ref{tab:special}, models fine-tuned on specific datasets, such as Watt-Tool-8B~\cite{watt-tool}, xLAM-7B~\cite{xlam}, and Hammer2.1-7B\cite{hammer}, exhibit a significant decline in performance on the Special dataset. This decline can primarily be attributed to the fact that while fine-tuning enhances a model’s performance on specialized tasks, it can also lead to a loss of generalization, making the model less effective on new or broader instruction-following tasks.This phenomenon highlights the importance of balancing task-specific performance and generalization capability during model optimization.

\noindent \textbf{Performance Limitations of Large Models in Complex Tasks.} As shown in Table~\ref{tab:agent_results}, most models exhibit an end accuracy of less than 50\% on Agent data tasks. This can be attributed to the fact that completing such tasks in dynamic environments, which simulate real-world multi-turn interactions, requires more than just performing individual tool operations. The model must also integrate contextual information during tool usage and account for the interdependencies between tool calls, which significantly increases task complexity. Furthermore, these tasks demand advanced reasoning and adaptability, which even large models may struggle with due to the challenges of maintaining consistency across long-term interactions and responding to the evolving nature of the task.

\begin{table}[htb]
\centering
\renewcommand{\arraystretch}{1} 
\small 
\caption{Performance evaluation of different models on Agent Data: \textbf{PA} represents Process Accuracy, \textbf{EA} represents End-to-End Accuracy (\%).}
\resizebox{\linewidth}{!}{ 
\begin{tabular}{lcccc}
\toprule
\multirow{2}{*}{\centering Model} & \multicolumn{2}{c}{\textbf{Multi Turn}} & \multicolumn{2}{c}{\textbf{Multi Step}} \\
\cmidrule(lr){2-3} \cmidrule(lr){4-5}
 & \textbf{EA} & \textbf{PA} & \textbf{EA} & \textbf{PA} \\
\midrule
GPT-4-Turbo & 50.0 & 66.0 & 85.0 & 89.5 \\
DeepSeek-V3 & 31.5 & 54.5 & 37.5 & 53.0\\
Claude-3-5-Sonnet & 21.5 & 41.5 & 57.5 & 76.5 \\
DouBao-Pro-32k & 20.0 & 45.5 & 30.0 & 47.5\\
Qwen2.5-7B-Instruct & 15.0 & 28.0 & 12.5 & 15.5 \\
Hammer2.1-7B & 8.5 & 33.5 & 25.0 & 42.5 \\
\bottomrule
\end{tabular}
}
\label{tab:agent_results}
\end{table}

\begin{table}[htbp]
\centering
\caption{Error type distribution across different model
series on Special Data.}
\label{tab:special_error_analysis} 
\resizebox{\linewidth}{!}{%
\begin{tabular}{lcc}
\toprule
\textbf{Model} & \textbf{Error Detection} & \textbf{Error Correction} \\
\midrule
Watt-Tool-8B & 188 & 4 \\
Hammer2.1-7B & 172 & 7 \\
Phi-3-mini-128k-instruct & 143 & 15 \\
Qwen2.5-3B-Instruct & 130 & 36 \\
xLAM-7B-r & 195 & 1 \\
Llama-3.1-8B-Instruct & 145 & 6 \\
Hammer2.1-3B & 197 & 0 \\
Llama-3.2-3B-Instruct & 166 & 9 \\
\bottomrule
\end{tabular}%
}
\end{table}

\subsection{Error Analysis}

\begin{figure}[t]
    \centering
    \includegraphics[width=0.95\columnwidth]{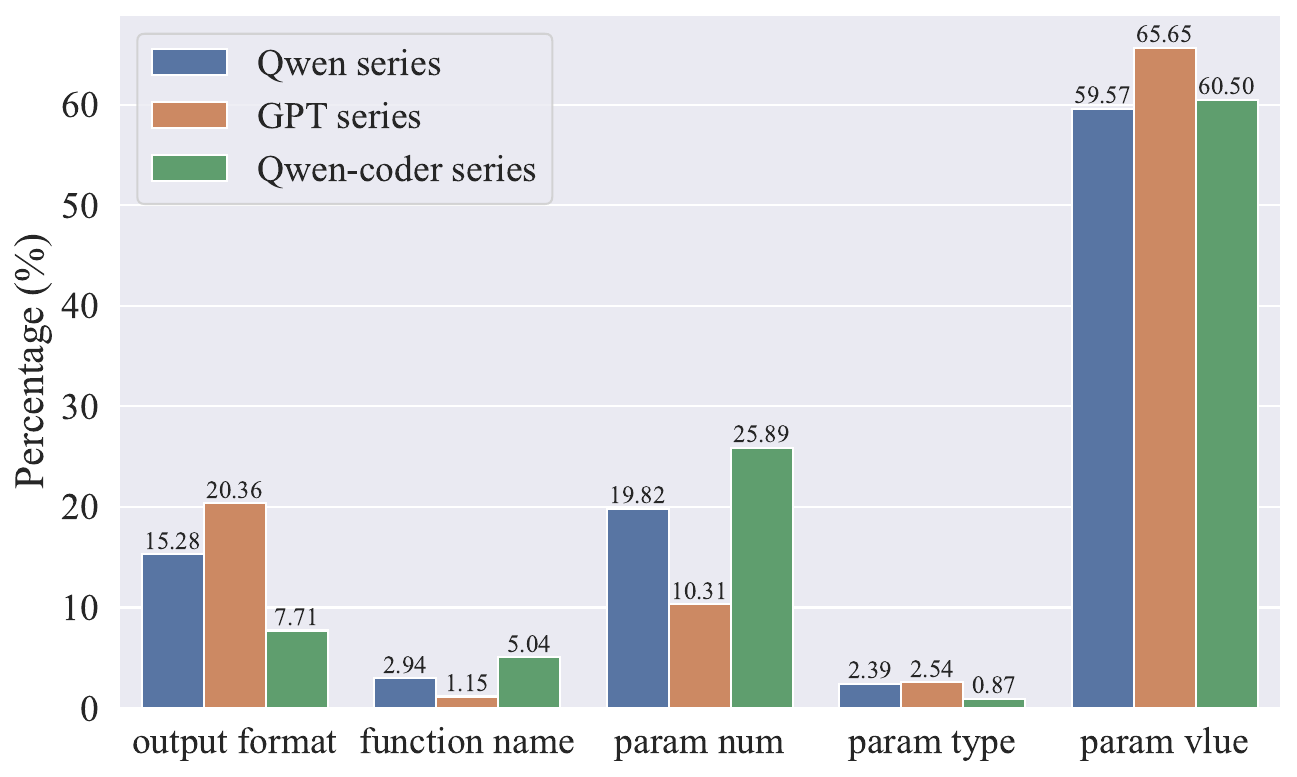} 
    \caption{Error type distribution on Normal Data.} 
    \label{fig:error_type} 
\end{figure}

 \textbf{Error Analysis of Normal Data.} As shown in Figure~\ref{fig:error_type}, we observe from the error type distribution on Normal data that param value error dominate across all models. This highlights the models' difficulty in generating specific values, likely due to limited contextual understanding and the complexity of numerical distributions. Output format error is the second most common, suggesting room for improvement in generating code that follows predefined formats and syntactic rules. These issues may stem from inconsistencies in training data and the models' limited ability to learn rule-based generation. In contrast, function name and param type errors are less frequent, indicating that the models excel in matching function calls and handling data types. While the models show strong function invocation abilities, further improvements are needed in numerical generation and format compliance. Specific error examples for Normal data can be found in Appendix~\ref{sec:normal_error}.

\noindent \textbf{Error Analysis of Special Data.}
As shown in Table~\ref{tab:special_error_analysis}, we identified two main types of model errors: The first type is "Error Detection", which refers to the model's complete failure to detect issues in the user's instructions or its inability to identify problems according to the prompt's formatting requirements.  The second type is "Error Correction," where the model detects the problem but provides unclear feedback. For example, the model might indicate that there is an issue, but fails to specify which parameter values are incorrect or what critical information is missing.
Results show that most errors in special-type scenarios are caused by "Error Detection", highlighting a critical gap in the model's problem-detection capabilities. This suggests that the model needs to learn not only simple tool invocation but also how to identify corresponding issues under imperfect instructions. Specific error examples can be found in Appendix~\ref{sec:special_error}.

\vspace{0.5em}
\noindent \textbf{Error Analysis of Agent Data.} 
Our analysis identifies three primary causes of Agent errors. First, function call errors occur when the model fails to select the appropriate function or provide parameters that do not meet the required specifications, reflecting a lack of understanding of tool-use capabilities and parameter constraints. Second, rule violations arise when the model disregards predefined scene rules, skipping necessary steps or breaking key task logic, highlighting deficiencies in its comprehension and execution. Finally, information mismanagement results from the model's inability to correctly record or process contextual information during multi-turn interactions, leading to outputs that diverge from expectations. As shown in Figure~\ref{fig:agent_example1}, we illustrate an error caused by missing information.

\subsection{Further Analysis}

\begin{figure}[t]
    \centering
    \includegraphics[width=0.8\linewidth]{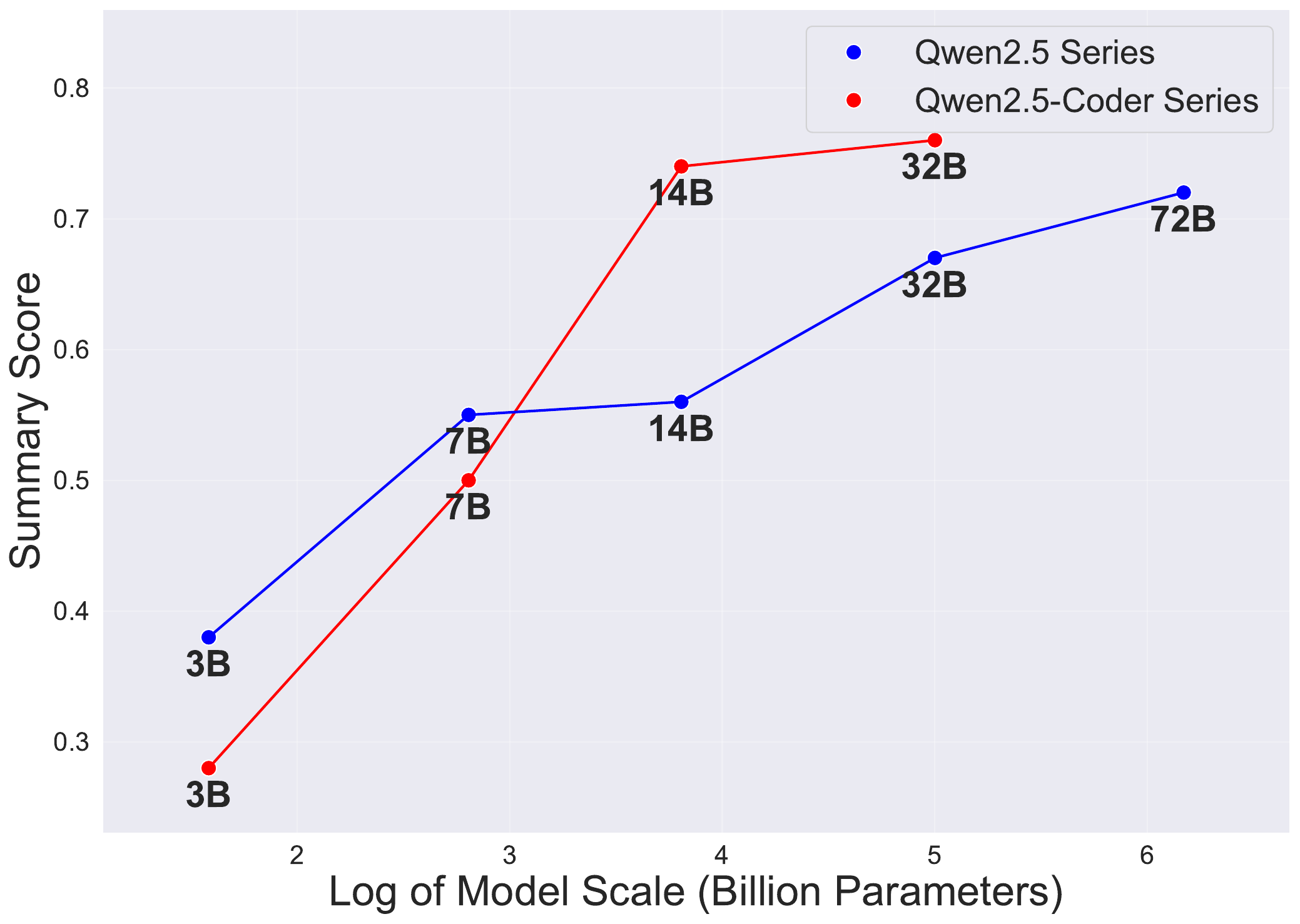} 
    \caption{Scaling Law of LLMs on ACEBench.} 
    \label{fig:scaling_law} 
\end{figure}

\textbf{Scaling Law.}
We evaluated the performance of Qwen2.5-Coder (3B, 7B, 14B, 32B) and Qwen2.5-Instruct (3B, 7B, 14B, 32B, 72B) on the ACEBench dataset. As shown in Figure~\ref{fig:scaling_law}, the experimental results demonstrate that performance improves significantly across various tasks as the model size increases, with particularly strong results observed in high-complexity tasks. However, it is worth noting that as the model size continues to grow, the rate of performance improvement begins to slow down, especially between the 32B and 72B models. This indicates that while increasing the model parameters brings substantial performance gains initially, the marginal benefits of scaling up further decrease, making additional improvements more challenging.

\begin{table}[htb]
\caption{Accuracy comparison of prompting strategies on English Normal Data (\%).}
\centering
\renewcommand{\arraystretch}{1} 
\resizebox{\linewidth}{!}{ 
\begin{tabular}{lcccc}
\toprule
\textbf{Model}  & \multicolumn{1}{c}{\textbf{Standard}} & \multicolumn{1}{c}{\textbf{Condensed}} & \multicolumn{1}{c}{\textbf{Minimal}} \\
\midrule
 Qwen2.5-3B-Instruct & 34.5 & 31.8 & 27.8  \\
 Qwen2.5-7B-Instruct & 48.5 & 47.5 & 45.5  \\
 Qwen2.5-14B-Instruct & 56.3 & 54.0 & 47.5  \\
\bottomrule
\end{tabular}
}
\label{tab:prompt}
\end{table}

 \noindent\textbf{Impact of Prompting Strategies.}
Prompt design significantly affects language model performance. We tested three strategies (see Appendix~\ref{sec:prompt_strategy}):

\noindent \textbf{(1)Standard Prompt:} A comprehensive template designed to eliminate interference from information insufficiency, ensuring a fair evaluation.

\noindent\textbf{(2)Condensed Prompt:} A compact version retaining core instructions, testing performance with reduced but sufficient guidance.

\noindent\textbf{(3)Minimal Prompt:} A highly abbreviated form (e.g keywords) to assess the model’s ability to infer tasks from ultra-concise input.

The experimental results in Table~\ref{tab:agent_results} demonstrate that models utilizing standard prompt templates achieve the highest overall accuracy. This optimal performance can be attributed to the rigorous formatting specifications in standard prompts, which effectively mitigate interference from extraneous variables. These empirical findings establish a positive correlation between prompt standardization and model performance, providing key insights for future prompt engineering: enhancing the standardization of function-calling prompts with explicit formatting requirements can significantly improve execution accuracy.

\section{Conclusion}

This paper introduces ACEBench, a comprehensive tool-use benchmark designed to evaluate the tool-use capabilities of Large Language Models (LLMs), including data from three types: normal, special, and agent. It addresses key limitations of existing evaluation benchmarks, such as the lack of multi-turn dialogue assessments in real-world scenarios, the absence of fine-grained evaluations for parameter-type function calls, and the high costs associated with using large models for evaluation.  The experimental results indicate that models fine-tuned on specific tool-use datasets to struggle with generalization when faced with complex or imperfect instructions, and code capabilities enhance the tool-use performance of large models. Through extensive experiments, we demonstrate the effectiveness of ACEBench in providing deeper insights into the tool-use abilities of various models.

\section*{Limitations}
We acknowledge several limitations in our evaluation of ACEBench for assessing the tool-use capabilities of large language models. Firstly, while our test data is generated by large language models and various measures have been taken to ensure its authenticity and diversity, a gap remains when compared to data from real-world applications. This discrepancy may impact the evaluation of the model's performance in real-world scenarios. Secondly, for the Agent data, the design of evaluation scenarios relies on manual construction, which somewhat limits the diversity and coverage of the evaluation framework.

\bibliography{main}

\appendix
\newpage 


\clearpage

\section{Detailed Descriptions of Test Cases}
\label{sec:data_description}
\subsection{Data Categories Description}
We divide the benchmark into three main categories: Normal, Special, and Agent. Below is a detailed description of each category.

\subsubsection*{Normal Data}
The Normal Data consists of fixed question-answer pairs, where each question corresponds to a correct function call. It is categorized into the following categories: Single-Turn, Multi-Turn, Similar APIs, Preference, and Atom. 

\textbf{Single-Turn}: There is only one round interaction between the user and the assistant, and based on the number of function calls in the response, it is divided into single-turn single function calls and single-turn parallel function calls.

\textbf{Multi-Turn}: There are multiple interactions between the user and the assistant. The conversation can be categorized into two types: \textbf{(switch)} The conversation progresses by changing topics.  \textbf{(adjust)} The conversation evolves by refining or modifying the original question.

\textbf{Similar APIs}: The candidate APIs exhibit significant similarity, particularly focusing on the same topic. This similarity presents a challenge for the assistant, requiring it to effectively distinguish between the APIs and accurately select the most appropriate ones.

\textbf{Preference}: Besides the candidate APIs, the assistant is provided with supplementary user profile data. This type of information necessitates the assistant's ability to mine user-specific factors, such as past interactions, interests, or other personalized attributes, to generate argument values.

\textbf{Atom}: Atom Data refers to a set of APIs that contain only specific parameter types, such as candidate functions where the parameters exclusively involve numbers, lists, etc. This design is intended to explore whether the type of function parameters affects the model's ability to handle data filling. We have divided the Atom data into five types: number, enum, list, bool, and object.

\subsubsection*{Special Data}
The Special Data refers to situations where the model is unable to resolve the problem posed in the instruction using the candidate functions\cite{nosiy_toolbench}. It is categorized into the following categories: Incomplete, Error, and Irrelevant.

\textbf{Incomplete}: Refers to situations where the key information required for the function call is missing in the query, such as the absence of "required" parameters.

\textbf{Error}: Refers to situations where the instruction contains parameters or names that do not meet the required format or constraints, such as matching a specific pattern or being selected from a predefined list, causing the function call to fail.

\textbf{Irrelevant}: Refers to situations where the instruction exceeds the function's capabilities, meaning none of the candidate functions can resolve the issue.

\subsubsection*{Agent Data}
Agent Data refers to scenarios where completing a task in an environment modeled after real-world situations typically requires multi-step collaboration. In this study, we employ the GPT-4o language model to simulate user roles and replicate real-world interaction processes, thereby evaluating the model’s performance in complex interactive settings. The key scenarios are defined as follows:

\textbf{Multi-step Scenario}: The user participates in only a single interaction throughout the entire dialogue flow.

\textbf{Multi-turn Scenario}: The user engages in multiple interactions across the entire dialogue cycle.

Agent Data currently encompasses the following fundamental scenarios:

\textbf{(1)Mobile Application Simulation.}
The mobile application scenario provides digital lifestyle functionalities including communication services, integrated reminder and memo management systems, and alarm configuration capabilities. This environment simulates core smartphone operations with particular attention to notification handling and scheduling precision.

\textbf{(2)Food Delivery Platform.}
This scenario simulates the core functionalities of a food delivery platform, primarily including merchant search, product browsing and ordering, order status tracking, and cancellation processing. The system implements essential operational procedures from merchant selection to order completion, supporting users throughout the entire food ordering experience.

\textbf{(3)Financial Services Scenario.}
This module provides fundamental banking service simulations, primarily including: deposit/withdrawal transactions, account balance inquiries, fund transfers, and other routine banking operations, while also supporting loan applications and repayment processes. The system maintains detailed transaction records and can generate basic financial statements, replicating the core services of real banking systems.

\textbf{(4)Travel Booking Platform.}
This scenario simulates a standard ticketing system, enabling users to complete end-to-end operations including flight/train ticket inquiry, booking, payment, rescheduling, and cancellation. The system incorporates fare checking, seat selection, and order management functionalities, capable of handling itinerary changes and related ticket adjustments. It covers the complete user journey from search to ticket issuance.

Our team is actively working on designing more functional scenarios to enhance the platformin Agent Data.

\begin{figure*}[t]
  \centering
  \includegraphics[width=\linewidth]{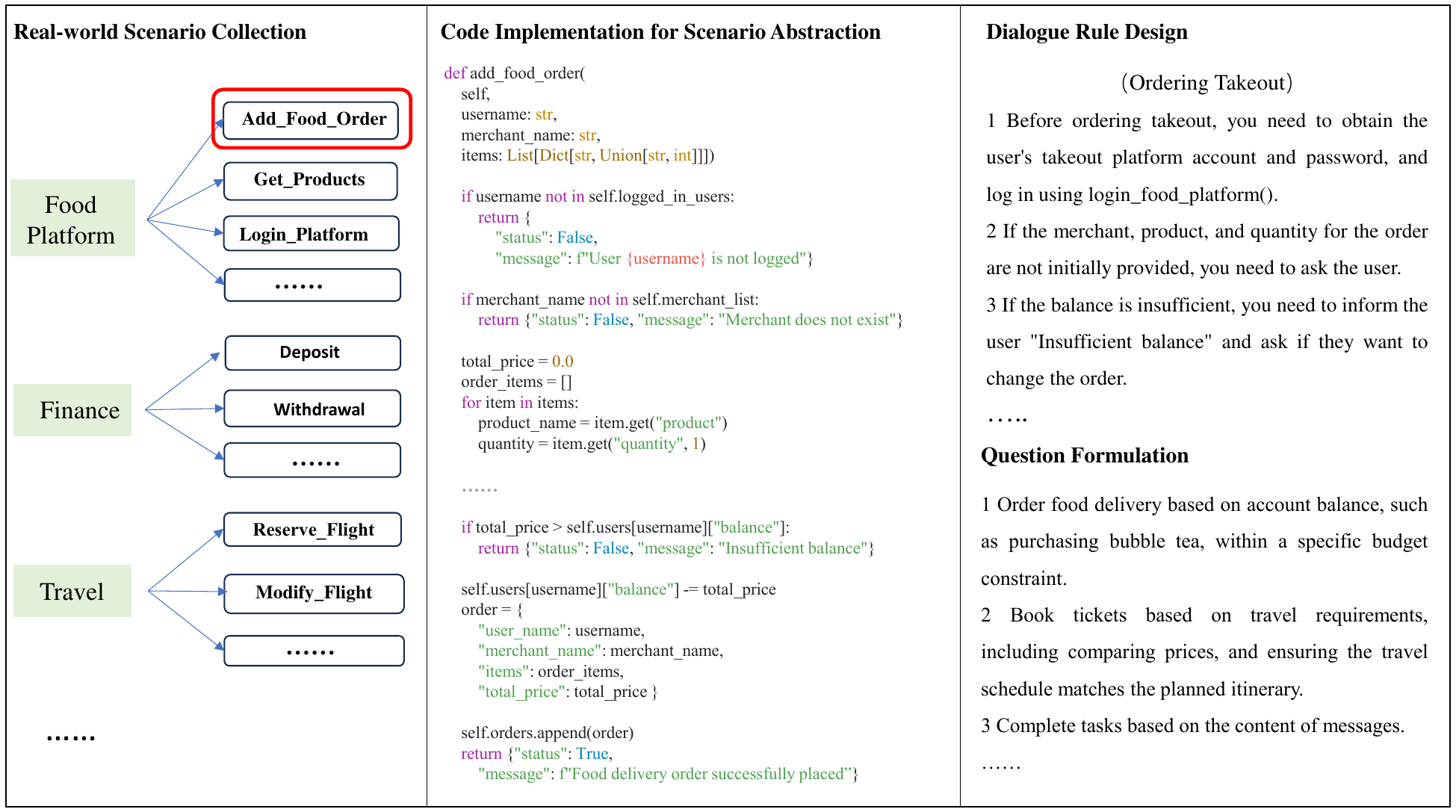}
  \caption{The construction of Agent Data. The left shows scenario sampling from real-world cases, the middle demonstrates the implementation of code tailored to specific scenarios, and the right presents examples of dialogue rules and question design for the scenarios.}
  \label{fig:agent_construct}
\end{figure*}

\section{The Construction of Data} 
\label{sec:construct_data}
\subsection{Agent Data} 
\label{sec:constructe_agent}
As shown in Figure~\ref{fig:agent_construct}, the construction of Agent data can be summarized in the following steps:
\vspace{0.5em}

First, through an in-depth analysis of real-world scenarios, extract key task requirements and modularize them into sub-scenarios in different domains (such as flight booking, food delivery platforms, and financial services), clearly define the specific functional objectives of each module. 

\vspace{0.5em}
Second, implement logical abstraction for each function through code, designing core processes such as user authentication, cost calculation, payment processing, and information recording. This ensures the code logic's scalability and robustness while comprehensively addressing exception handling. 
\vspace{0.5em}

Third, design interaction rules based on specific scenario requirements to standardize the interaction process between users and the Agent, such as verifying account, merchant information, and balance status in food delivery orders, and providing user guidance in exceptional cases. 
\vspace{0.5em}

Finally, combine real-world demands to design question formats and solutions, enabling the model to accurately meet user task requirements and achieve the desired outcomes.

\subsection{Special Data}
\label{sec:construct_special}

\subsubsection*{Irrelevant}
The irrelevant data refers to situations where the instruction exceeds the function’s capabilities. And the construction method for Irrelevant data is similar to that of Normal data, where we simply remove the correct API from the candidate APIs of the common data.

\subsubsection*{Incomplete}
The special data with incomplete instructions primarily refers to situations where key information is missing from the user's instructions, causing the function to be called incorrectly. Our main prompt for constructing incomplete data is shown in Figure~\ref{fig:incomplete_construct}. From the generated dialogue, we can extract the corresponding data and reference answers (the missing data). Next, we can convert the conversation we obtained into data. A specific example is shown in Figure~\ref{fig:incomplete_construct_example}.

\begin{figure}[H]
\centering
\small
\begin{mdframed}
Please refer to the example and continue the dialogue based on the given tool definition and the beginning of the conversation. \textbf{The requirements are as follows:}

1 The user's request is meaningful, requiring the use of one tool, and the tool will be called once.

2 Calling the tool to fulfill the user's request still lacks one or more required parameters.

......

Here is an example for reference\\

\textbf{\texttt{<tool\_definition>}} 
\begin{verbatim}
{
"name": "calculate_triangle_area", 
"description": "Given the base and height of a 
                triangle, calculate its area.", 
"required": ["base", "height"]
}
\end{verbatim}

\textbf{\texttt{<dialogue\_example>}} 
\begin{verbatim}
[User]: I want to calculate the area of a 
triangle with a base of 5 and a height of 10.
[Assistant]: 
[Thought]  Missing parameters: base|height  
[Response]:  
Please provide both the base and the 
height of the triangle.
[User]: The base is 5, and the height is 10.
[Assistant]: 
<tool_usage>calculate_triangle_area
|{"base": 5, "height": 10}</tool_usage>
\end{verbatim}
\textbf{Here is the dialogue continuation you need to write: }

<Tool Definition> 

\{tool\_definition\}

......

\end{mdframed}
\caption{Main prompt of Incomplete Data construction.}
\label{fig:incomplete_construct}
\end{figure}

\begin{figure}[H]
\centering
\small
\begin{mdframed}
\textbf{candidate function}
\begin{verbatim}
{
  "name": "book_flight",
  "description": "Flight booking ...",
  "arguments": {},
  "required": ["from_city_name", 
   "to_city_name", "depart_date"]
}
\end{verbatim}
\textbf{obtained dialouge}
\begin{verbatim}
<user> I plan to travel to Beijing.
Help me book a flight to Beijing.
</user> [Thought] Missing parameters:
from_city_name 
[Response] May I know the departure 
city you want to book?
<user> Shenzhen
\end{verbatim}

\textbf{converted data} 
\begin{verbatim}
[question] I plan to travel to Beijing.
Help me book a flight to Beijing.
[answer] Miss necessary parameter 
(from_city_name) from (book_flight)
\end{verbatim}
\end{mdframed}
\caption{Specific example of Incomplete Data construction.}
\label{fig:incomplete_construct_example}
\end{figure}

\begin{figure}[H]
\centering
\small
\begin{mdframed}

Please refer to the example and continue the dialogue based on the given tool definition and the beginning of the conversation. \textbf{The requirements are as follows:}

1 The user's request is specific and meaningful, requiring one tool, and the tool should be called once.

2 The user's first question contains incorrect parameters (i.e., it does not conform to the function definition’s pattern or format requirements).

......

Here is an example for reference

\textbf{\texttt{<tool\_definition>}} 
\begin{verbatim}
{"name": "TennisTeamDetails",
"description": "Retrieve detailed 
information about a tennis team .",
"parameters": {
   "properties": {
     "team_name": {......
    "pattern": "^[a-zA-Z\\s]+$"}}}}

\end{verbatim}

\textbf{\texttt{<dialogue\_example>}} 
\begin{verbatim}
<user> Can you retrieve the details of 
     the tennis team named 'Team@1234'?
</user> [Thought] Error Parameter: 'Team@1234'
[Response] The team name you provided does
          not meet the naming criteria.
<user> Oh, I made a mistake. 
   The team name is TeamABC.
</user> <tool_usage>TennisTeamDetails
|{"team_name": "TeamABC"}</tool_usage>"
\end{verbatim}
\textbf{Here is the dialogue continuation you need to write: }

<Tool Definition> 

\{tool\_definition\}

......

\end{mdframed}
\caption{Main prompt for Error Data construction.}
\label{fig:special_error_construct}
\end{figure}

\subsubsection*{Error}
The special data which has error instructions mainly refers to situations where the instruction contains parameters or names that do not meet the required format or constraints, and the construction is shown in~\ref{fig:special_error_construct}. Next, we can convert the conversation we obtained into data. A specific example is shown in~\ref{fig:error_construct_example}.

\begin{figure}[H]
\centering
\small
\begin{mdframed}
\textbf{candidate function}
\begin{verbatim}
{ "name": "FootballTeamDetails",
  "description": "Retrieve information 
  about a football team by its name.",
  "parameters": {
      "team_name": {......
        "pattern": "^[a-zA-Z\\s]+$"}}}}
\end{verbatim}
\textbf{obtained dialouge}
\begin{verbatim}
<user> I want to know information about 
the football team football$156.
</user> [Thought] Error Parameter:
         football$156.
[Response] The team name you provided 
does not meet the naming criteria.
<user> ......
\end{verbatim}

\textbf{converted data} 
\begin{verbatim}
[question] I want to know information 
about the football team footbaoo$156.
[answer] There is incorrect value
(football$156) for the (team_name).
\end{verbatim}
\end{mdframed}
\caption{Specific example of Error Data construction.}
\label{fig:error_construct_example}
\end{figure}

\section{Examples of Dataset}
\label{sec:data_example}
\subsection{Normal Examples} 

\textbf{Single-Turn.} The example of Normal Single-Turn Data is shown in Figure~\ref{fig:normal_single_turn}.

\noindent\textbf{Multi-Turn.} The example of Normal Multi-Turn Data is shown in Figure~\ref{fig:normal_multi_turn}.

\noindent\textbf{Preference.} The example of Normal Preference Data is shown in Figure~\ref{fig:normal_profile}.

\noindent\textbf{Similar APIs.} The example of Normal Similar APIs Data is shown in Figure~\ref{fig:similar_api}.

\noindent\textbf{Atom.} The example of Normal Atom Data is shown in Figure~\ref{fig:atom_enum}.

\subsection{Special Examples} 

\textbf{Incomplete.} The example of Special Incomplete Data is shown in Figure~\ref{fig:special_Incpmplete}.

\noindent\textbf{Error.} The example of Special Error Data is shown in Figure~\ref{fig:special_error_param}.

\noindent\textbf{Irrelevant.} The example of Special Irrelevant Data is shown in Figure~\ref{fig:special_irrelevant}.

\subsection{Agent Examples} 
\begin{figure*}[ht]
  \centering
  \includegraphics[width=\textwidth]{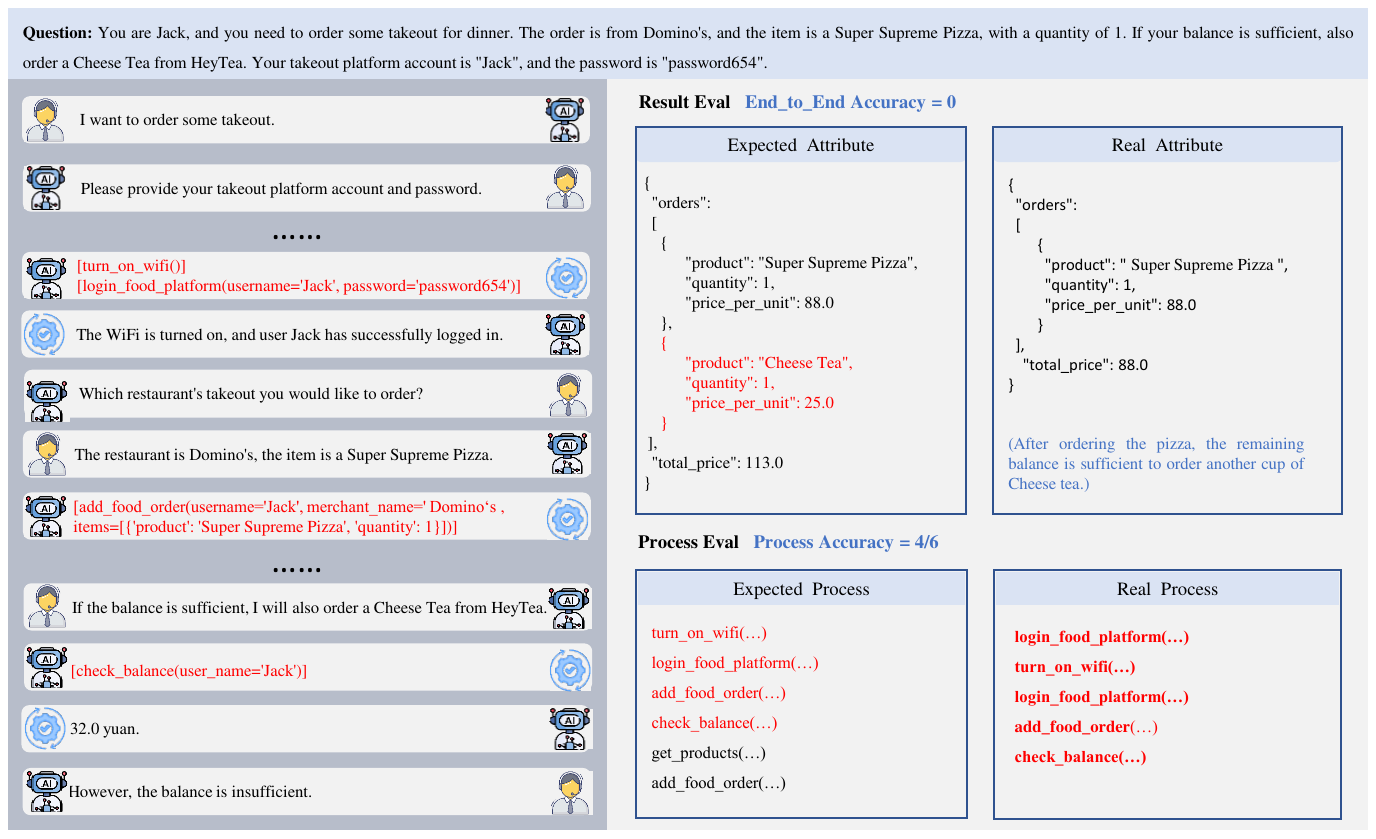}
  \caption{An example of Agent Data: A partially correct representation.}
  \label{fig:agent_example1}
\end{figure*}

\begin{figure*}[ht]
  \centering
  \includegraphics[width=\textwidth]{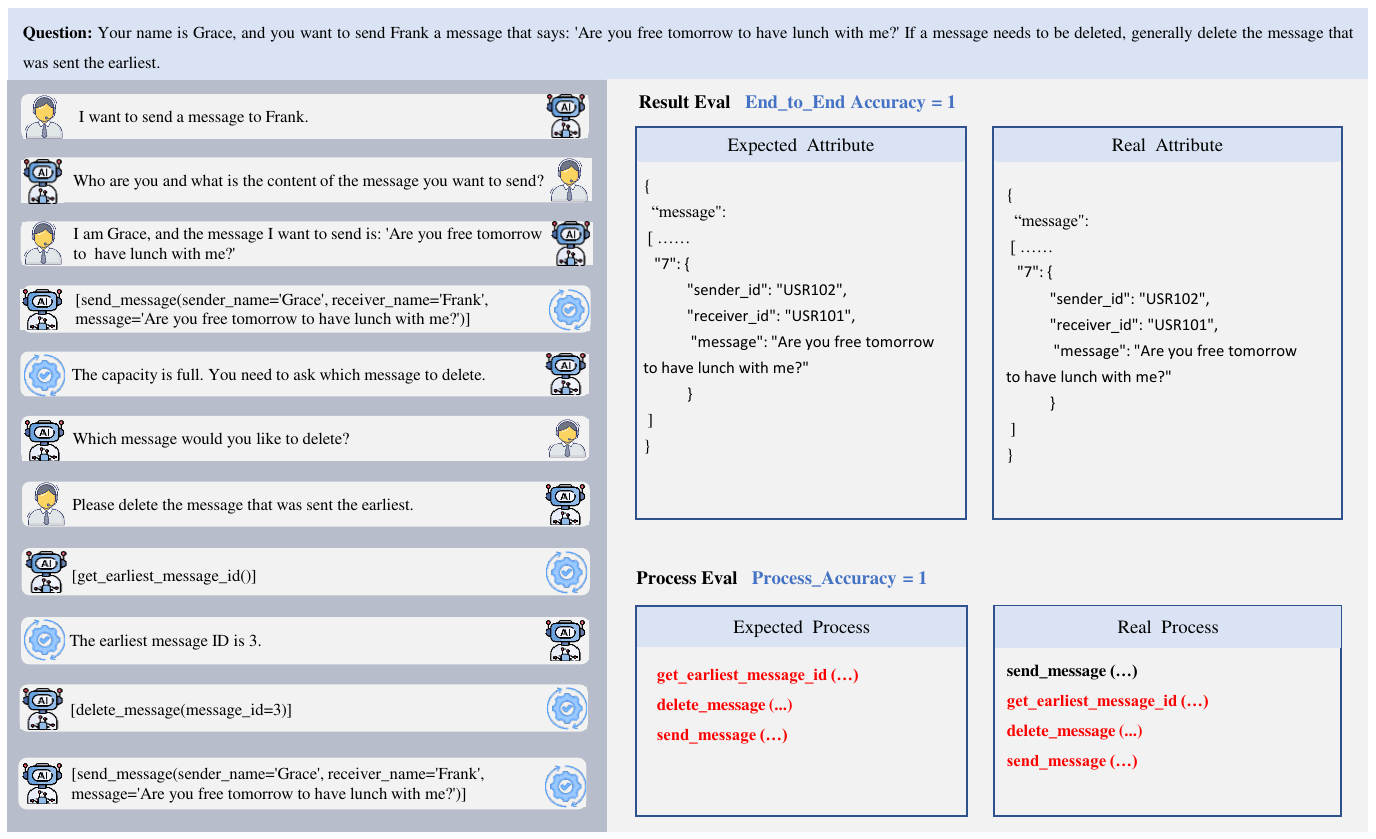}
  \caption{An example of Agent Data: A fully accurate representation.}
  \label{fig:agent_example2}
\end{figure*}

The example of Agent Data is shown in Figure~\ref{fig:agent_example1} and Figure~\ref{fig:agent_example2}.

\section{Evaluation Inference Prompts}
\label{sec:inference_prompt}
\subsection{Normal Prompt} 
\label{sec:normal_prompt}
The main evaluation inference prompt for Normal Data is shown in Figure~\ref{fig:normal_prompt_0} and Figure~\ref{fig:normal_prompt_1}.

\subsection{Special Prompt} 
\label{sec:special_prompt}
The main evaluation inference prompt for Special Data is shown in Figure~\ref{fig:special_prompt}.

\subsection{Agent Prompt} 
\label{sec:agent_prompt}
An example of the evaluation inference prompt for Agent Data in a specific scenario is shown in Figure~\ref{fig:agent_prompt}. And an inference prompt is shown in Figure~\ref{fig:user_prompt}.

\subsection{Different Prompt Strategies} 
\label{sec:prompt_strategy}
Standard Prompt is shown in Figure~\ref{fig:normal_prompt_0}.
Condensed Prompt is shown in Figure~\ref{fig:easy_prompt}.
Minimal Prompt is shown in Figure~\ref{fig:mini_prompt}.

\section{Formula for Overall Accuracy} 
\label{sec:formula}
The formula for calculating the \textbf{Overall Accuracy} can be expressed as:

\[
\text{All Acc} = A \cdot \text{Acc}_{\text{Normal}} + B \cdot \text{Acc}_{\text{Special}} + C \cdot \text{Acc}_{\text{Agent}}
\]

where the coefficients \(A\), \(B\), and \(C\) are defined as:

\[
A = \frac{\sqrt{n_{\text{Normal}}}}{\sqrt{n_{\text{Normal}}} + \sqrt{n_{\text{Special}}} + \sqrt{n_{\text{Agent}}}}
\]

\[
B = \frac{\sqrt{n_{\text{Special}}}}{\sqrt{n_{\text{Normal}}} + \sqrt{n_{\text{Special}}} + \sqrt{n_{\text{Agent}}}}
\]

\[
C = \frac{\sqrt{n_{\text{Agent}}}}{\sqrt{n_{\text{Normal}}} + \sqrt{n_{\text{Special}}} + \sqrt{n_{\text{Agent}}}}
\]

where \(n_{\text{Normal}}, n_{\text{Special}}, n_{\text{Agent}}\) are the sample sizes for the \textbf{Normal}, \textbf{Special}, and \textbf{Agent} data types, and \( \text{Accuracy} \) refers to the accuracy of each respective category.

\section{Detailed Results}
\label{sec:detailed_results}
The accuracy evaluation of different models on English Data is shown in Table~\ref{tab:en_results}, and the accuracy evaluation of different models on Chinese Data is shown in Table~\ref{tab:zh_results}.

\section{Error Examples} 
\subsection{Error Examples of Normal Data}
\label{sec:normal_error}

\textbf{Wrong Function Name.} An example of wrong function name is shown in Figure~\ref{fig:error_normal_1}.

\noindent\textbf{Wrong Param Type.} An example of wrong param type is shown in Figure~\ref{fig:error_normal_2}.

\noindent\textbf{Wrong Param Value.} An example of wrong param value is shown in Figure~\ref{fig:error_normal_3}.

\noindent\textbf{Wrong Output Format.} An example of wrong output format is shown in Figure~\ref{fig:error_normal_4}.


\subsection{Error Examples of Special Data}
\label{sec:special_error}
\textbf{Error Detection}
Figure~\ref{fig:error_Special_1} shows an example of error detection.

\noindent\textbf{Error Correction}
Figure~\ref{fig:error_Special_2} shows an example of error correction.

\onecolumn
\begin{figure}[H]
    \centering
    \small
    \begin{mdframed}
    \textbf{(question)} \\\\
    Please assess the system with asset ID 12345, vulnerability ID 67890, with a severity level of high, and a potential threat that may lead to data leakage.\\

    \textbf{(candidate function)}
    \begin{verbatim}
        {
        "name": "Threat_Vulnerability_Assessment_Tool",
        "description": "Assess potential security 
        threats and provide mitigation suggestions",
        "parameters": ...... 
        }
        {
        "name": "userPrivacySettingsManager",
        "description": "Manage user's privacy",
        "parameters": ...... 
        }
    \end{verbatim}
    \end{mdframed}
    \caption{An example of Normal Single-Turn Data.}
    \label{fig:normal_single_turn}
\end{figure}


\begin{figure}[H]
    \centering
    \small
\begin{mdframed}
\textbf{(conversation)} \\ \\
\noindent \textbf{User:} I want to analyze the pest and disease risks faced by my rice crop, which is currently in the tillering stage.\\
\noindent \textbf{system:} Please provide the current temperature and humidity conditions.user: The temperature is 28°C, and the humidity is 80\%.\\
\noindent \textbf{tool:}
\begin{verbatim}
[PestInfestationRiskAnalyzer(
    crop_information={"crop_type": "rice", "growth_stage": "tillering"}, 
    climate_conditions={"temperature": 28, "humidity": 80})]
\end{verbatim}
\noindent \textbf{system:} Based on the analysis, your rice is facing a moderate risk of pests and diseases under the current climatic conditions. Please take necessary preventive measures.\\
\noindent \textbf{User:} I also want to know the maintenance requirements for my production equipment. The device ID is Device123.\\
\noindent \textbf{system:} Please provide the sensor readings of the device, including temperature, vibration, and humidity.\\
\noindent \textbf{User:} The device currently reads a temperature of 75°C, vibration level of 5, and surrounding humidity of 50\%.\\

\textbf{(candidate function)}
\begin{verbatim}
{
  "name": "PestInfestationRiskAnalyzer",
  "description": "Analyze the pest and disease risk faced by the farmland ",
  "parameters": ...... 
}
{
  "name": "Predictive_Maintenance_AI_Tool",
  "description": "Predicting the maintenance needs of equipment using sensor data",
  "parameters": ...... 
}
{
  "name": "PersonalityTraitAnalyzer",
  "description": "Analyze an individual's personality traits",
  "parameters": ...... 
}
\end{verbatim}

\end{mdframed}
    \caption{An example of Normal Multi-Turn Data.}
    \label{fig:normal_multi_turn}
\end{figure}


\begin{figure}[H]
    \centering
    \small
\begin{mdframed}
\textbf{(question)} \\
Can you update my preferred notification method and also check if my current email address and home location are properly updated in the system?\\

\textbf{(candidate function)}
\begin{verbatim}
{
  "name": "updateOrderStatusAlerts",
  "description": "Sends automated alerts to users regarding the status of their current orders.",
  "parameters": ...... 
}
{
  "name": "submitProductReview",
  "description": "Allows users to submit a review for a product they have purchased.",
  "parameters": ...... 
}
{
  "name": "updateUserProfile",
  "description": "Updates the user's profile information based on provided data.",
  "parameters": ...... 
}
\end{verbatim}

\textbf{(profile)}
\begin{verbatim}
{
  "basic_features": {
    "UserName": "Michael Smith",
    "UserEmail": "mike.smith@example.com",
    "UserHomeLocation": "Los Angeles, CA",
    "UserBirthday": "1978-04-23",
    "UserLanguage": "Spanish",
    "UserTimeZone": "PST",
    ....... 
  },
  "user_history": {
    "shopping": [
      "Searched for 'Nike running shoes' on app",
      "Added Nike Air Max to cart",
      "Checked coupon availability for Nike products",
      "Filtered search by 'Outdoor Equipment' category", 
      "Selected 'High spending' filter for items over $500",
      ...... 
    ],
    "takeout": [
      "Ordered Chicken Fajitas on the takeout app for lunch",
      "Opted to receive promotional deals via phone calls",
      "Chose Debit Card ending in 5678 for payment on the takeout app",
      ...... 
    ]
  }
}
\end{verbatim}

\end{mdframed}
    \caption{An example of Normal Preference Data.}
    \label{fig:normal_profile}
\end{figure}


\begin{figure}[H]
    \centering
    \small
\begin{mdframed}
\textbf{(question)} \\\\
My baby has had a visible vein on her nose for 5 days, and she's been crying a lot with a decreased appetite. Can you help?\\

\textbf{(candidate function)}
\begin{verbatim}
{
  "name": "baby_health_check_A",
  "description": "Checks the common reasons for baby's persistent vein visibility on the nose 
      and suggests actions. This API considers factors like skin thinness, crying, or overexertion",
  "parameters": ...... 
}
{
  "name": "baby_health_check_B",
  "description": "Examines baby's vein visibility and recommends seeing a doctor. 
      Focuses on persistent visibility and associated symptoms",
  "parameters": ...... 
}
\end{verbatim}
\end{mdframed}
    \caption{An example of Similar API Data.}
    \label{fig:similar_api}
\end{figure}

\begin{figure}[H]
    \centering
    \small
\begin{mdframed}
\textbf{(question)} \\\\
I need a design for my new website. It's for a technology company focusing on user engagement.\\

\textbf{(candidate function)}
\begin{verbatim}
{
  "name": "WebDesignAssistant_generateDesign",
  "description": "Generates a website design based on industry and user experience focus.",
  "parameters": {
    "type": "object",
    "properties": {
      "industry": {
        "description": "The industry for which the website is being designed.",
        "type": "string",
        "enum": [
          "Technology",
          "Healthcare",
          "Education",
          "Finance"
        ],
        "default": "Technology"
      },
      "userExperience": {......}
    },
    "required": ["industry", "userExperience"]
  }
}
{……}
\end{verbatim}
\end{mdframed}
    \caption{An example of Atom (enum) Data.}
    \label{fig:atom_enum}
\end{figure}


\begin{figure}[H]
    \centering
    \small
\begin{mdframed}
\textbf{(question)} \\\\
I'm considering relocating my business to the Middle East. Can you provide me with a list of major cities?\\

\textbf{(candidate function)}
\begin{verbatim}
{"name": "Get_Middle_East_Cities",
  "description": "Retrieves a list of cities in the Middle East, sorted by overall score by default.",
  "parameters": {
    "properties": {
      "sort": {
        "description": "The sorting order for the list of cities.",
        "type": "string",
        "enum": ["asc", "desc"],
      }
    },
    "required": ["sort"]
  }}    
\end{verbatim}
\end{mdframed}
    \caption{An example of Special Incomplete Data.}
    \label{fig:special_Incpmplete}
\end{figure}

\begin{figure}[htb]
    \centering
    \small
\begin{mdframed}
\textbf{(question)} \\\\
Can you retrieve the tennis team details named 'Team@1234’?”\\

\textbf{(candidate function)}
\begin{verbatim}
{"name": "TennisTeamDetails",
  "description": "Retrieve detailed information about a tennis team by its name.",
  "parameters": {
    "type": "object", 
    "properties": {
      "team_name": {
        "pattern": "^[a-zA-Z\\s]+$", 
      }
    },
    "required": ["team_name"]
  }}
\end{verbatim}
\end{mdframed}
    \caption{An example of Special Error Param Data.}
    \label{fig:special_error_param}
\end{figure}

\begin{figure}[H]
    \centering
    \small
\begin{mdframed}
\textbf{(question)} \\\\
Could you help me find available restaurants in New York City?\\

\textbf{(candidate function)}
\begin{verbatim}
{
  "name": "Get_Weather_Report",
  "description": "Retrieve the current weather report for a specified location",
  "parameters": ...... 
}
{
  "name": "GetTravelDestinationInfo",
  "description": "Retrieves information about a specific travel destination",
  "parameters": ...... 
}
\end{verbatim}
\end{mdframed}
    \caption{An example of Special Irrelevant Data.}
    \label{fig:special_irrelevant}
\end{figure}

%

\clearpage

\begin{figure}[htb]
    \centering
    \small
\begin{mdframed}
You are an AI assistant with the role name "assistant". Based on the provided API specifications and conversation history from steps 1 to t, generate the API requests that the assistant should call in step t+1. The API requests should be output in the format [ApiName(key1='value1', key2='value2', ...)], replacing ApiName with the actual API name, key1, key2, etc., with the actual parameter names, and value1, value2, etc., with the actual parameter values. The output should start with a square bracket "[" and end with a square bracket "]".
If there are multiple API requests, separate them with commas, for example:[ApiName(key1='value1',key2='value2',...),ApiName(key1='value1',key2='value2', ...), ...]. Do not include any other explanations, prompts, or API call results in the output.
If the API parameter description does not specify otherwise, the parameter is optional (parameters mentioned in the user input need to be included in the output; if not mentioned, they do not need to be included).
If the API parameter description does not specify the required format for the value, use the user's original text for the parameter value.
If the API requires no parameters, output the API request directly in the format [ApiName()], and do not invent any nonexistent parameter names.\\

\textbf{\{time\}}\\\\
\textbf{Role Descriptions:}\\
user: User \\
assistant: The AI assistant role that makes API requests\\
tool: Provides the results returned from tool calls\\
\\
\textbf{API Specifications}:\\
\{function\}

\end{mdframed}
    \caption{The inference prompt for Nomal (except Prefernce) Data.}
    \label{fig:normal_prompt_0}
\end{figure}

\begin{figure}[htb]
    \centering
    \small
\begin{mdframed}
You are an AI assistant, and your role is called assistant. \textbf{Based on the given API description, dialogue history 1..t, and character profile}, generate the API requests that the assistant should call in step t+1. The API requests should be output in the format [ApiName(key1='value1', key2='value2', ...)], where ApiName is replaced with the actual API name, and key1, key2, etc., are replaced with the actual parameter names, and value1, value2 are replaced with the actual parameter values. The output should start with a "[" and end with a "]".
If there are multiple API requests, they should be separated by commas, e.g., [ApiName(key1='value1', key2='value2', ...), ApiName(key1='value1', key2='value2', ...), ...]. Do not output any other explanations, hints, or results of the API calls in the output.
If the API parameter description does not specify special instructions, the parameter is optional (parameters mentioned in the user input or character profile should be included in the output, and if not mentioned, they should not be included).
If the API parameter description does not specify the format for the parameter value, it should be taken from the user's original text or character profile.
If the API requires no parameters, the API request should be output as [ApiName()], with no fabricated parameter names.\\

Character Profile:\\
\textbf{\{profile\}}\\\\
\textbf{Role Descriptions:}\\
user: User \\
assistant: The AI assistant role that makes API requests\\
tool: Provides the results returned from tool calls\\
\\
\textbf{API Specifications}:\\
\{function\}

\end{mdframed}
    \caption{The inference prompt for Nomal (Prefernce) Data.}
    \label{fig:normal_prompt_1}
\end{figure}

\begin{figure}[htb]
    \centering
    \small
\begin{mdframed}
You are an AI assistant with the role name "assistant". Based on the provided API specifications and conversation history from steps 1 to t, generate the API requests that the assistant should call in step t+1. The API requests should be output in the format [ApiName(key1='value1', key2='value2', ...)], replacing ApiName with the actual API name, key1, key2, etc., with the actual parameter names, and value1, value2, etc., with the actual parameter values. The output should start with a square bracket "[" and end with a square bracket "]".\\

\textbf{\{time\}}\\\\
\textbf{Role Descriptions:}\\
user: User \\
assistant: The AI assistant role that makes API requests\\
tool: Provides the results returned from tool calls\\
\\
\textbf{API Specifications}:\\
\{function\}

\end{mdframed}
    \caption{Condensed Prompt for Normal Data.}
    \label{fig:easy_prompt}
\end{figure}

\begin{figure}[htb]
    \centering
    \small
\begin{mdframed}
You are an AI assistant. Based on the provided API specifications and conversation history generate the API requests in the format
[ApiName(key1='value1', key2='value2', ...),.....].\\

\textbf{\{time\}}\\\\
\textbf{Role Descriptions:}\\
user: User \\
assistant: The AI assistant role that makes API requests\\
tool: Provides the results returned from tool calls\\
\\
\textbf{API Specifications}:\\
\{function\}

\end{mdframed}
    \caption{Minimal Prompt for Normal Data.}
    \label{fig:mini_prompt}
\end{figure}

\begin{figure}[htb]
    \centering
    \small
\begin{mdframed}
You are an AI system with the role name "assistant". Based on the provided API specifications and
conversation history from steps 1 to t, generate the API requests that the system should call in step t+1.
Below are two specific scenarios:\\
\textbf{1. When the information provided by the user is clear and unambiguous, and the problem can be
resolved using the list of candidate functions:}\\
- If the API parameter description does not specify the required format for the value, use the user's
original text for the parameter value.\\
- API requests should be output in the format [ApiName(key1='value1', key2='value2', ...),
ApiName(key1='value1', key2='value2', ...), ...], replacing ApiName with the actual API name, key1, key2,
etc., with the actual parameter names, and value1, value2, etc., with the actual parameter values. The
output should start with a square bracket "[" and end with a square bracket "]". At this time, the output
must not contain any other content.\\
\textbf{2. When the information provided by the user is unclear, incomplete, or incorrect, or the user's
question exceeds the capabilities of the provided functions, you need to clearly point out these issues.
The following is your strategy:}\\
(1) If the user's instructions include the key details required to call the API, but the type or form of the
parameter values does not match the API's definitions, ask in-depth questions to clarify and correct the
details. The output format should be: ["There is incorrect value (value) for the parameters (key) in the
conversation history."]\\
(2) If the user's instructions lack the key details required by the API, ask questions to obtain the
necessary information. The output format should be: ["Missing necessary parameters (key1, key2, ...) for
the api (ApiName)"], replacing key1, key2 with the names of the missing parameters and ApiName with
the actual API name.\\
(3) If the user's request exceeds the current capabilities of your APIs, inform them that you cannot fulfill
the request. The output format should be: ["Due to the limitations of the function, I cannot solve this
problem."]\\
\textbf{Note: The above steps have a priority order. You need to first determine whether scenario (1)
applies. If it does, output according to the requirements in (1). Pay attention to distinguishing
between scenarios (1) and (2).}\\

\textbf{\{time\}}\\\\
\textbf{Role Descriptions:}\\
user: User \\
assistant: The AI assistant role that makes API requests\\
tool: Provides the results returned from tool calls\\
\\
\textbf{API Specifications}:\\
\{function\}

\end{mdframed}
    \caption{The inference prompt for Special Data.}
    \label{fig:special_prompt}
\end{figure}

\begin{figure}[ht]
    \centering
    \small
\begin{mdframed}
The current time is June 11, 2024, 16:00 (Beijing Time). As a simulated mobile assistant agent, you can help users \textbf{send text messages, add reminders, and order takeout.}\\\\
\textbf{Text messages}\\
\textbf{Sending Text Messages} (1)Before sending a text message, the agent must first obtain the sender and recipient of the message.(2)When the memory is full and needs to delete messages, you need to ask the user: "Memory is full, which message would you like to delete?"\\
\textbf{Viewing Text Messages} (1)Before viewing text messages, the agent must first log into the device via login\_device().(2)Before viewing text messages, the agent must first obtain the sender and recipient of the messages.(3)After viewing text messages, the agent needs to ask the user if they want to add the message content to a reminder.(4)After viewing text messages, the agent needs to ask the user if they want to reply to the message.(5)If the message content involves takeout, the agent needs to ask if the user wants to order takeout based on the message content.\\\\
\textbf{Reminders}\\
\textbf{Adding Reminders}(1)Before adding a reminder, you should obtain the content and title of the reminder. The reminder time defaults to the current time.(2)If the reminder to be added is the content of a specific message, the agent needs to first view the message content.\\
\textbf{Viewing Specific Reminders by Title}:After viewing a specific reminder by title, you need to ask the user if they want to complete the tasks within it.\\\\
\textbf{Order takeout}\\
\textbf{Ordering Takeout}(1)Before ordering takeout, the agent needs to obtain the user's takeout platform account and password, and log in using login\_food\_platform().(2)If the merchant, product, and quantity for the order are not initially provided, you need to ask the user.(3)When encountering takeout from different merchants, you need to order them one by one.(4)If the balance is insufficient, you need to inform the user "Insufficient balance" and ask if they want to change the order.\\\\
\textbf{Function Calls}\\
When a function call is needed, please strictly adhere to the above format requirements:\\
 (1)[ApiName(key1='value1', key2='value2', ...)], Please remember that the function call must start with [ and end with ]\\
(2)You need to promptly feedback the task execution status to the user and do not repeatedly call the same function. When you believe the current task is completed, respond with "finish conversation" to end the dialogue.\\

\end{mdframed}
    \caption{The inference prompt for Agent Data in a specific scenario.}
    \label{fig:agent_prompt}
\end{figure}

\clearpage

\begin{figure}[b]
    \centering
    \small
\begin{mdframed}
As a user, your role is to interact with an agent. However, during the interaction, you need to follow these guidelines:

1 Break down your inquiries and only raise one question per exchange to simulate a real user's messages.

2 Provide all the necessary information for the current step. For instance, when setting a reminder, you must give details such as the reminder's description, title, and time.

3 When asked if you require further assistance, ensure that the main tasks in the instruction have been completed. If not, continue to present the next step to the agent.

4 When the agent asks which message needs to be deleted, proceed with the deletion as specified in the instructions. You cannot offer proactive help to the agent; respond to the agent's questions according to the instructions, and do not invent any information that you do not know.

5 Once all tasks are complete, generate a 'finish conversation' message as a standalone line to end the discussion.

Question: {question}

\end{mdframed}
    \caption{An inference prompt for user simulator in a specific scenario.}
    \label{fig:user_prompt}
\end{figure}

\clearpage

\newpage

\begin{table}[H]
    \centering
    \caption{Accuracy evaluation of different models on English Data (\%).}
    \resizebox{\textwidth}{!}{%
    \begin{tabular}{lcccccccccc}
        \toprule
        \multirow{2}{*}{\textbf{Model}} & 
        \multicolumn{6}{c}{\textbf{Normal}} & 
        \multicolumn{1}{c}{\multirow{2}{*}{\textbf{Special}}} & 
        \multicolumn{1}{c}{\multirow{2}{*}{\textbf{Agent}}} & 
        \multicolumn{1}{c}{\multirow{2}{*}{\textbf{Overall}}} \\
        \cmidrule(lr){2-7}
        & \textbf{Atom} & \textbf{Single-Turn} & \textbf{Multi-Turn} & \textbf{Similar API} & \textbf{Preference} & \textbf{Summary} & & & \\
        \midrule
        \multicolumn{10}{c}{\textbf{Closed-Source Large Language Models}} \\
        \midrule
        GPT-4o & 90.0 & 78.0 & 68.0 & 80.0 & 78.0 & 82.5 & 92.7 & 56.0 & \textbf{81.1} \\
        GPT-4-Turbo & 90.7 & 80.5 & 69.0 & 80.0 & 88.0 & 84.2 & 82.0 & 62.5 & \textbf{80.3} \\
        Qwen-Max & 88.0 & 75.0 & 61.0 & 74.0 & 82.0 & 79.7 & 74.0 & 60.0 & \textbf{75.1} \\
        GPT-4o-Mini & 84.3 & 73.5 & 59.0 & 74.0 & 72.0 & 76.4 & 76.7 & 27.5 & \textbf{68.9} \\
        Gemini-1.5-Pro & 82.3 & 73.0 & 61.0 & 74.0 & 72.0 & 75.7 & 77.3 & 26.0 & \textbf{68.5} \\
        Claude-3-5-Sonnet & 66.7 & 64.0 & 46.0 & 58.0 & 68.0 & 62.2 & 72.7 & 44.0 & \textbf{62.2} \\
        Doubao-Pro-32k & 75.3 & 58.0 & 52.0 & 70.0 & 54.0 & 66.3 & 50.7 & 26.5 & \textbf{56.0} \\
        \midrule
        \multicolumn{10}{c}{\textbf{Open-Source Large Language Models}} \\
        \midrule
        Kimi-k2-0711 & 87.0 & 78.5 & 62.0 & 70.0 & 74.0 & 78.9 & 81.3 & 65.0 & \textbf{77.4} \\
        Qwen2.5-Coder-32B-Instruct & 86.0 & 73.5 & 59.0 & 76.0 & 72.0 & 77.4 & 80.0 & 50.0 & \textbf{73.9} \\
        LoopTool-8B & 86.0 & 76.0 & 58.0 & 74.0 & 78.0 &78.0 & 80.7 & 43.3 & \textbf{73.4}\\
        DeepSeek-V3 & 88.0 & 77.5 & 63.0 & 76.0 & 78.0 & 80.3 & 72.7 & 34.0 & \textbf{71.1} \\
        Qwen2.5-72B-Instruct & 81.3 & 74.5 & 64.0 & 76.0 & 80.0 & 76.8 & 74.0 & 37.5 & \textbf{70.0} \\
        Llama-3.1-70B-Instruct & 83.7 & 71.5 & 61.0 & 74.0 & 66.0 & 75.6 & 29.3 & 41.0 & \textbf{57.9} \\
        Qwen2.5-7B-Instruct & 70.3 & 57.0 & 49.0 & 62.0 & 58.0 & 62.8 & 49.3 & 15.0 & \textbf{51.8} \\
        Qwen2.5-Coder-7B-Instruct & 73.3 & 63.5 & 52.0 & 70.0 & 58.0 & 66.6 & 25.3 & 18.5 & \textbf{48.1} \\
        DeepSeek-Coder-V2-Lite-Instruct & 71.7 & 58.0 & 50.0 & 62.0 & 60.0 & 64.0 & 39.3 & 2.5 & \textbf{47.9} \\
        Watt-Tool-8B & 84.7 & 71.5 & 57.0 & 70.0 & 62.0 & 74.8 & 2.0 & 1.5 & \textbf{44.0} \\
        Hammer2.1-7B & 71.3 & 62.5 & 43.0 & 64.0 & 52.0 & 62.9 & 3.3 & 15.0 & \textbf{39.6} \\
        Phi-3-Mini-128k-Instruct & 66.3 & 49.0 & 31.0 & 58.0 & 32.0 & 54.0 & 12.0 & 0.0 & \textbf{34.4} \\
        MLlama-3.1-8B-Instruct & 51.0 & 49.5 & 28.0 & 60.0 & 56.0 & 48.1 & 15.3 & 6.5 & \textbf{32.9} \\
        xLAM-7B-r & 61.7 & 42.0 & 32.0 & 66.0 & 0.0 & 48.7 & 4.0 & 10.0 & \textbf{30.8} \\
        Llama-3.2-3B-Instruct & 31.7 & 21.5 & 9.0 & 34.0 & 32.0 & 26.4 & 8.7 & 0.0 & \textbf{17.6} \\
        Hammer2.1-3B & 32.7 & 14.0 & 7.0 & 36.0 & 32.0 & 25.5 & 0.7 & 1.5 & \textbf{15.2} \\

        \bottomrule
    \end{tabular}%
    }
    \label{tab:en_results}
\end{table}

\begin{table}[H]
    \centering
    \caption{Accuracy evaluation of different models on Chinese Data (\%).}
    \resizebox{\textwidth}{!}{%
    \begin{tabular}{lcccccccccc}
        \toprule
        \multirow{2}{*}{\textbf{Model}} & 
        \multicolumn{6}{c}{\textbf{Normal}} & 
        \multicolumn{1}{c}{\multirow{2}{*}{\textbf{Special}}} & 
        \multicolumn{1}{c}{\multirow{2}{*}{\textbf{Agent}}} & 
        \multicolumn{1}{c}{\multirow{2}{*}{\textbf{Overall}}} \\
        \cmidrule(lr){2-7}
        & \textbf{Atom} & \textbf{Single-Turn} & \textbf{Multi-Turn} & \textbf{Similar API} & \textbf{Preference} & \textbf{Summary} & & & \\
        \midrule
        \multicolumn{10}{c}{\textbf{Closed-Source Large Language Models}} \\
        \midrule
        GPT-4o & 96.7 & 91.0 & 86.0 & 90.0 & 88.0 & 92.7 & 93.3 & 71.5 & \textbf{89.6} \\
        GPT-4-Turbo & 95.7 & 89.0 & 86.0 & 92.0 & 84.0 & 91.7 & 91.3 & 72.5 & \textbf{88.6} \\
        Qwen-Max & 94.3 & 86.0 & 75.0 & 92.0 & 84.0 & 88.7 & 74.0 & 68.5 & \textbf{81.7} \\
        GPT-4o-Mini & 88.7 & 78.5 & 74.0 & 80.0 & 84.0 & 83.4 & 81.3 & 39.0 & \textbf{76.0} \\
        Claude-3-5-Sonnet & 87.0 & 81.0 & 79.0 & 84.0 & 76.0 & 83.5 & 82.0 & 35.0 & \textbf{75.6} \\
        Gemini-1.5-Pro & 86.7 & 80.5 & 68.0 & 86.0 & 84.0 & 82.2 & 80.0 & 25.0 & \textbf{72.8} \\
        Doubao-Pro-32k & 84.3 & 53.0 & 64.0 & 82.0 & 78.0 & 75.0 & 59.3 & 23.5 & \textbf{62.8} \\
        \midrule
        \multicolumn{10}{c}{\textbf{Open-Source Large Language Models}} \\
        \midrule
        Qwen2.5-Coder-32B-Instruct & 94.3 & 88.5 & 83.0 & 90.0 & 90.0 & 90.8 & 81.3 & 71.5 & \textbf{85.3} \\
        Qwen2.5-72B-Instruct & 92.3 & 86.0 & 75.0 & 90.0 & 82.0 & 87.3 & 77.3 & 52.5 & \textbf{79.3} \\
        DeepSeek-V3 & 95.0 & 90.5 & 91.0 & 90.0 & 88.0 & 92.6 & 73.3 & 35.0 & \textbf{78.5} \\
        Llama-3.1-70B-Instruct & 81.3 & 65.0 & 66.0 & 84.0 & 70.0 & 75.3 & 47.3 & 43.5 & \textbf{62.9} \\
        Qwen2.5-7B-Instruct & 81.7 & 63.5 & 68.0 & 82.0 & 76.0 & 75.9 & 44.7 & 12.5 & \textbf{57.8} \\
        DeepSeek-Coder-V2-Lite-Instruct & 78.7 & 57.5 & 43.0 & 82.0 & 70.0 & 68.8 & 41.3 & 1.5 & \textbf{51.1} \\
        Qwen2.5-Coder-7B-Instruct & 78.7 & 64.0 & 63.0 & 78.0 & 78.0 & 73.5 & 19.3 & 12.5 & \textbf{49.6} \\
        Watt-Tool-8B & 86.7 & 67.0 & 54.0 & 88.0 & 66.0 & 76.3 & 10.0 & 4.0 & \textbf{47.4} \\
        Hammer2.1-7B & 76.0 & 62.5 & 37.0 & 60.0 & 58.0 & 62.7 & 26.0 & 18.5 & \textbf{46.1} \\
        Llama-3.1-8B-Instruct & 52.7 & 30.0 & 28.0 & 72.0 & 36.0 & 45.0 & 26.7 & 4.0 & \textbf{33.8} \\
        Phi-3-Mini-128k-Instruct & 48.0 & 29.5 & 15.0 & 58.0 & 32.0 & 38.9 & 25.3 & 1.5 & \textbf{29.5} \\
        Llama-3.2-3B-Instruct & 45.7 & 9.0 & 9.0 & 50.0 & 32.0 & 32.7 & 10.0 & 0.0 & \textbf{21.6} \\
        xLAM-7B-r & 25.3 & 2.0 & 6.0 & 56.0 & 0.0 & 18.7 & 1.3 & 7.5 & \textbf{12.3} \\
        Hammer2.1-3B & 12.0 & 9.0 & 0.0 & 44.0 & 8.0 & 11.8 & 1.3 & 1.5 & \textbf{7.4} \\
        \bottomrule
    \end{tabular}%
    }
    \label{tab:zh_results}
\end{table}

\begin{figure}[htb]
    \centering
    \small
\begin{mdframed}
\textbf{(question)} \\
I want to understand the symmetry in Escher's \"Waterfall\" artwork. Please provide a detailed symmetry analysis.\\\\
\textbf{(answer)} 
\begin{verbatim}
{"Escher_Artwork_Analysis_Tool": {
    "artwork_analysis": [
      {"artwork_id": "Waterfall",
       "analysis_type": "Symmetry Analysis"}]}}
\end{verbatim}
\textbf{(model output)}\\
\{\textcolor{red}{“FontAdvisor\_recommendDisplayFonts”}:\{......\}\}\\
\end{mdframed}
    \caption{An error example of Wrong Function Name on Normal Data.}
    \label{fig:error_normal_1}
\end{figure}

\begin{figure}[htb]
    \centering
    \small
\begin{mdframed}
\textbf{(question)} \\
Can you check if our current automation system with system ID 2435 is compatible with the new device model XR10?\\\\
\textbf{(answer)} \\
\{"AutomationSystemCompatibilityCheck":\{ "systemID":\textcolor{red}{2435},"newDeviceModel":"XR10"\}\}\\\\
\textbf{(model output)} \\
\{“AutomationSystemCompatibilityCheck”:\{ “systemID”:\textcolor{red}{"2435"}, "newDeviceModel":"XR10"\}\}
\end{mdframed}
    \caption{An error example of Wrong Param Type on Normal Data.}
    \label{fig:error_normal_2}
\end{figure}

\begin{figure}[htb]
    \centering
    \small
\begin{mdframed}
\textbf{(question)} \\
I would like to assess the GPS accuracy of my vehicle. The device ID is GPS-7524. Sorry, I made a mistake earlier—it should be GPS-7526.\\\\
\textbf{(answer)} \\
\{ "SignalAccuracyEvaluator":\{ "device\_id":\textcolor{red}{"GPS-7526“} \}\}
\\\\
\textbf{(model output)} \\
\{"SignalAccuracyEvaluator":\{ "device\_id":\textcolor{red}{"GPS-7524“}\}\}
\end{mdframed}
    \caption{An error example of Wrong Param Value on Normal Data.}
    \label{fig:error_normal_3}
\end{figure}

\begin{figure}[H]
    \centering
    \small
\begin{mdframed}
\textbf{(question)} \\
I am currently working on a research paper about the state of forests, and I need an assessment of the vegetation diversity in the Amazon rainforest. Please provide me with a detailed report.\\\\
\textbf{(answer)} 
\begin{verbatim}
{"EcoDiversityMonitor_assessVegetation": {
    "area_identifier": "Amazon rainforest",
    "detailed_report": true}}
\end{verbatim}
\textbf{(model output)}\\
\begin{verbatim}
[user,area_identifier='Amazonrainforest',detailed_report=True]
\end{verbatim}
\end{mdframed}
    \caption{An error example of Wrong Output Format on Normal Data.}
    \label{fig:error_normal_4}
\end{figure}

\clearpage

\begin{figure}[htb]
    \centering
    \small
\begin{mdframed}
\textbf{(question)} \\
I‘m considering relocating my business to the Middle East. Can you provide me with a detailed list of major cities there, sorted by internet speed?\\\\
\textbf{(answer)} \\
Missing necessary parameters (sort) for the API (Get\_Middle\_East\_Cities)\\\\
\textbf{(model output)} 
\begin{verbatim}
{
  "Get_Middle_East_Cities": {
    "sort": "desc",
    "sort_by": "internet_speed",
    "size": 10,
    "page": 1
  }
}
\end{verbatim}
\textcolor{red}{(The question did not mention the sort method)}
\end{mdframed}
    \caption{An error example of Error Detection on Special Data.}
    \label{fig:error_Special_1}
\end{figure}

\begin{figure}[htb]
    \centering
    \small
\begin{mdframed}
\textbf{(question)} \\
I‘m considering relocating my business to the Middle East. Can you provide me with a detailed list of major cities there, sorted by internet speed?\\\\
\textbf{(answer)} \\
Missing necessary parameters (sort) for the api (Get\_Middle\_East\_Cities)\\\\
\textbf{(model output)} \\
Missing necessary parameters (sort by) for the api (Get\_Middle\_East\_Cities)\\
\textcolor{red}{(Find missing necessary parameters, but not correctly identified)}
\end{mdframed}
    \caption{An error example of Error Correction on Special Data.}
    \label{fig:error_Special_2}
\end{figure}


\end{document}